\documentclass[11pt]{article}

\usepackage[final]{acl}

\usepackage{times}
\usepackage{latexsym}

\usepackage[T1]{fontenc}

\usepackage[utf8]{inputenc}

\usepackage{microtype}

\usepackage{inconsolata}

\usepackage{graphicx}

\title{Solve the Missing First Step: Can VLMs Standardize \\ Raw Heterogeneous Medical Data?}

\author{
  \textbf{Xin Chen},
  \textbf{Dongliang Xu\textsuperscript{*}},
  \textbf{Cunhao Zhu},
  \textbf{Xudong Luo},
  \textbf{Haoyang Lyu},
  \\
  \textbf{Xiaoxiao Sun},
  \textbf{Serena Yeung-Levy},
  \textbf{Yue Yao\textsuperscript{*}}
\\
}

\usepackage{booktabs}
\usepackage{tabularx}
\usepackage{array}
\usepackage{longtable}
\usepackage{booktabs}
\usepackage{array}

\newcommand{\ie}{\textit{i.e.}}
\usepackage{booktabs}
\usepackage{amsmath}
\usepackage{amssymb}
\usepackage{bbm}
\usepackage{multirow}
\usepackage{graphicx}
\usepackage{multirow}
\usepackage[table]{xcolor}

\definecolor{bestgreen}{RGB}{166, 217, 155}
\definecolor{secondgreen}{RGB}{226, 245, 218}

\newcommand{\best}[1]{\cellcolor{bestgreen}\textbf{#1}}
\newcommand{\second}[1]{\cellcolor{secondgreen}#1}

\begin{document}
\maketitle

\vspace{-2mm}

\begingroup
\renewcommand{\thefootnote}{*}
\footnotetext{Corresponding authors. Code: \href{https://github.com/chenxin-sdu/MDS-Bench}{MDS-Bench}}
\endgroup
\vspace{-1mm}

\begin{abstract}
As vision-language models (VLMs) are increasingly applied to medical AI, existing benchmarks mainly focus on evaluating their diagnostic ability over given medical images and texts, implicitly assuming that standardized medical images, texts, or question–answer pairs are already prepared. However, this assumption does not hold when we apply VLMs in real clinical practice, where medical data is often raw, heterogeneous, and fragmented across different sources. In this paper, we study this missing step, \ie, raw medical data standardization. Specifically, models are given raw dataset folders and evaluated on their ability to identify source formats, convert raw medical images into VLM-compatible visual inputs, extract relevant textual information, and organize the results into structured image-text pairs. To construct this Medical Data Standardization Benchmark (MDS-Bench), we manually annotate 1,939 raw medical data standardization tasks covering diverse clinical practice, radiology modalities, annotation formats, and directory layouts. Extensive experiments show that even the best performing VLM, \ie, Gemini 3 Flash, achieves only a 48.6\% end-to-end success rate. Our research highlights raw medical data standardization as a critical bottleneck for medical AI diagnosis in real practice.

\end{abstract}

\section{Introduction}

\begin{figure}[t]
    \centering
    \includegraphics[width=\linewidth]{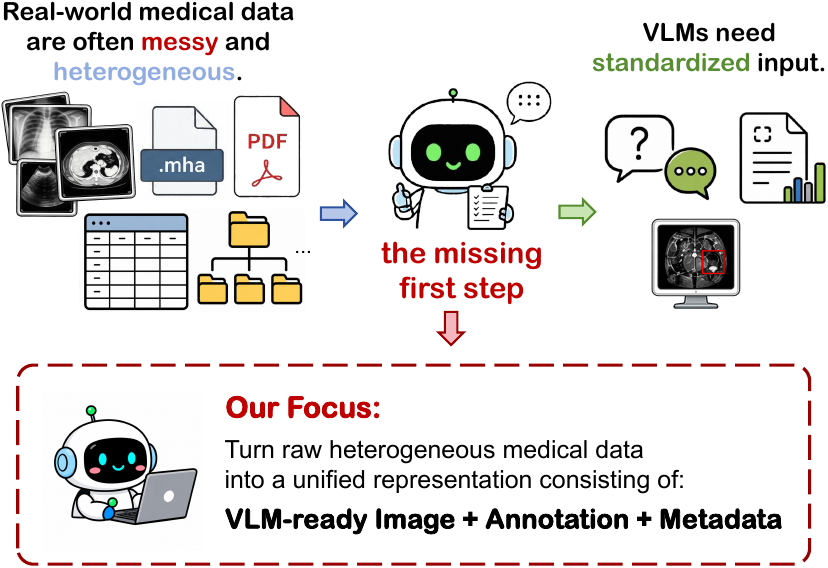}
    \caption{Motivation for raw medical data standardization. 
    When medical AI systems are deployed in the real world, their inputs are not always standardized or VLM-ready. Instead of receiving well-prepared images, reports, or question–answer pairs, VLMs may need to process raw clinical data collected from different sources, where relevant files, annotations, metadata, and directory layouts are not explicitly aligned. Therefore, an upstream standardization step is needed to convert raw medical inputs into representations that can be directly used by medical VLMs.
    }
    \label{fig:motivation}
\end{figure}

\begin{figure*}[t]
    \centering
    \includegraphics[width=1\linewidth]{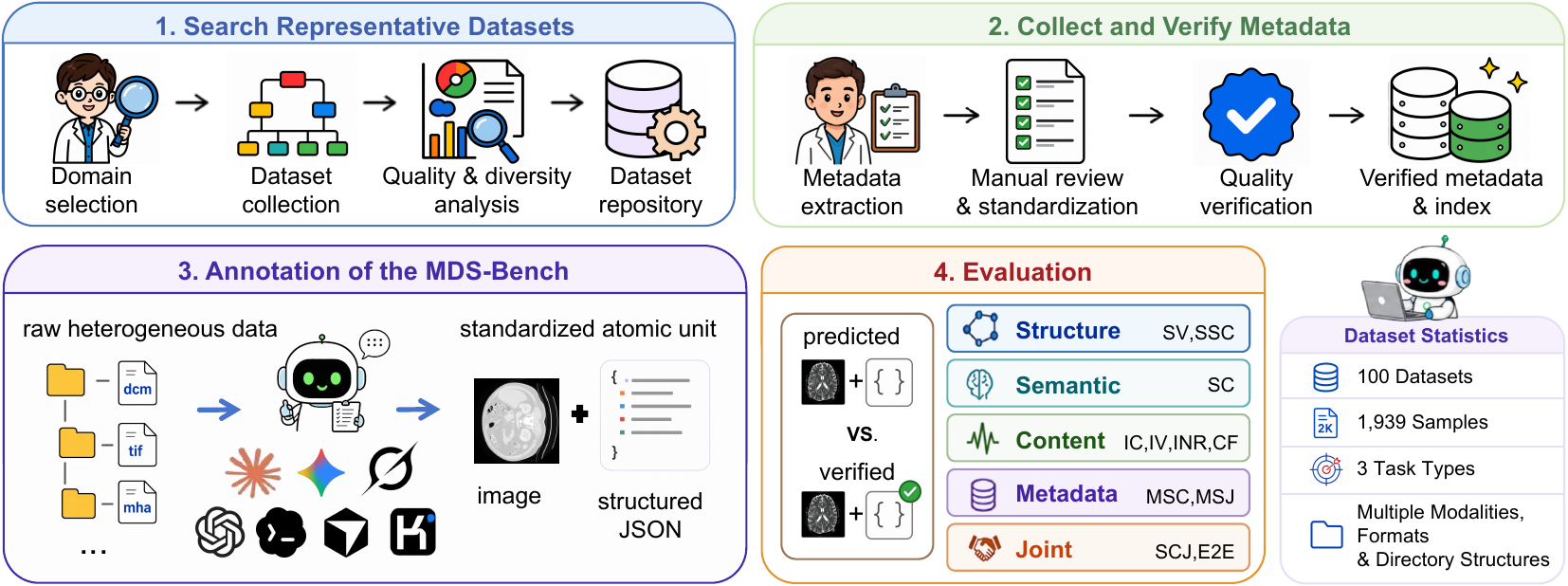}
    \caption{Overview of the MDS-Bench construction pipeline. 
    We collect representative raw medical imaging datasets, trace target samples to their original visual sources, annotations, and metadata, and verify standardized image-JSON outputs against the raw resources. The resulting benchmark covers diverse task types, modalities, raw file formats, annotation styles, and directory structures.
    }
    \label{fig:method}
\end{figure*}

With the rapid progress of VLMs, building intelligent systems that can directly process real-world medical data is becoming increasingly feasible~\cite{li2023llava}. However, existing medical benchmarks usually start from curated images, reports, or question-answer pairs~\cite{chen2024gmai}. This design is useful for evaluating perception and reasoning after the input has been cleaned, but it leaves this upstream standardization problem largely unexplored. In practical medical workflows, data are often stored as heterogeneous files, annotations, metadata records, masks, volumes, and dataset specific folders~\cite{hosseinzadeh2025data,pezoulas2019medical,barret2025etl,zhang2022heterogeneous,liu2025review,deng2026project}.  Before a VLM can reason over these resources, it should first identify the relevant sources, align them, convert them, and organize them into a usable representation~\cite{wilkinson2016fair}. This missing step limits our understanding of how well VLMs can operate on real medical data. Figure~\ref{fig:motivation} illustrates this gap: real-world medical resources should first be identified, aligned, and standardized before they can serve as VLM-ready image-text inputs.

Raw medical data standardization is challenging because medical datasets rarely follow a single organization rule. A single sample may depend on multiple resources, such as a raw image, a segmentation mask, a label table, a metadata record, or a slice from a volumetric scan~\cite{wasserthal2023totalsegmentator}. The links among these resources are often implicit in file names, directory structures, annotation conventions, or dataset documentation. Thus, standardization is not merely a format-conversion problem: a model must recover sample structure, select the correct sources, align annotations, and preserve task-relevant medical semantics. If this step fails, downstream inputs may appear complete while still being grounded in the wrong source, annotation, or clinical meaning.

\begin{figure*}[t]
    \centering
    \includegraphics[width=\linewidth]{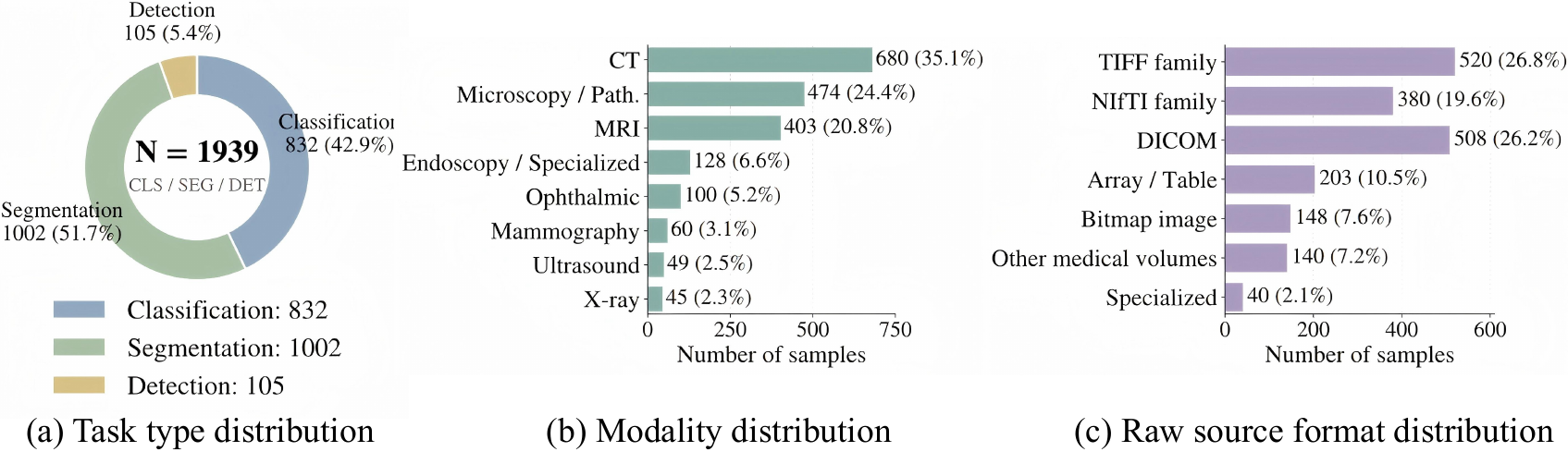}
    \caption{Dataset composition of MDS-Bench. The benchmark spans classification, segmentation, and detection tasks, and includes diverse medical imaging modalities and heterogeneous raw source formats.}
    \label{fig:benchmark_stats}
\end{figure*}

We study this missing step as raw medical data standardization in an agentic VLM setting, where a VLM is embedded in a coding-agent environment and can interact with files, tools, and executable scripts~\cite{yao2022react,akella2025data,li2025autodcworkflow}. Specifically, in this task, a model receives a raw dataset folder. We expect our VLMs can identify source files,  convert raw medical images into VLM-compatible visual inputs, align annotations and metadata, and generate a unified image and JSON output. The image provides a common visual representation for downstream models, while the JSON records information that cannot be captured by the image alone, such as source references, modality attributes, task type, labels, masks, boxes, and dataset metadata. This task differs from medical image classification, visual question answering, and report generation, which usually assume that the relevant input has already been selected. Our task starts earlier and asks whether VLMs can make raw medical resources usable while preserving their original meaning.

We build MDS-Bench to make raw medical data standardization measurable. As shown in Figure~\ref{fig:method}, the benchmark is constructed by collecting raw medical imaging datasets, tracing target samples to their original sources, and verifying standardized outputs against source evidence. It contains 1,939 samples across classification, segmentation, and detection, covering diverse modalities, file formats, annotation styles, and directory layouts. Since a model may produce valid JSON while choosing the wrong source or misaligning annotations, we further introduce an eleven-metric evaluation protocol that assesses image-JSON structure, sample-level semantics and fidelity, dataset-level metadata quality, and strict full-pipeline success. 

Extensive experiments show that current VLMs still fail to standardize raw medical data reliably. For example, 
even the best performing VLM, \ie, Gemini 3 Flash, achieves only a 48.6\% end-to-end success rate, indicating that this upstream task remains far from solved. 
The main difficulty is not schema generation alone, but preserving source based semantics, content fidelity, metadata consistency, and full-pipeline correctness across heterogeneous raw resources. 
We provide a systematic empirical study showing that raw data standardization is a major bottleneck for applying agentic multimodal models to real medical data.

\section{Related Work}

\textbf{Existing Medical AI Benchmarks.}
Existing medical AI benchmarks mainly evaluate perception, question answering, diagnosis, and reasoning after inputs have already been curated.
Representative benchmarks, including MultiMedQA~\cite{singhal2023large}, BioASQ~\cite{nentidis2025overview}, PMC-VQA~\cite{zhang2023pmc}, GMAI-MMBench~\cite{chen2024gmai}, MMMU~\cite{yue2024mmmu}, and MMBU~\cite{d2026mmbu}, measure medical knowledge, biomedical question answering, multimodal understanding, or expert-level reasoning.
These benchmarks have advanced medical AI evaluation, but they usually start from selected images, organized question-answer pairs, or predefined annotations.
Recent workflow-oriented systems, such as MedAgents~\cite{tang2024medagents}, MedPerf~\cite{karargyris2023federated}, and MedAgentBench~\cite{jiang2025medagentbench}, move closer to real deployment by studying agent collaboration and practical evaluation pipelines.
However, they view data preparation as a prerequisite instead of an evaluative capability.
Our work benchmarks this missing upstream step: whether VLMs can recover usable samples from raw heterogeneous medical resources, align visual content with annotations and metadata, and produce source-grounded image-text outputs for downstream use.

\noindent\textbf{Structured Output and Multimodal Structure Extraction.}
Structured output generation and multimodal structure extraction study whether VLMs can produce outputs with explicit schemas.
In text settings, StructEval~\cite{yang2025structeval} and JSONSchemaBench~\cite{geng2025jsonschemabench} evaluate schema following for language models.
In multimodal settings, Image2Struct~\cite{roberts2024image2struct} and SO-Bench~\cite{feng2025so} evaluate structured generation from visual inputs.
These works show that schema compliance is important for VLM reliability, but their inputs are usually already prepared, such as clean images, text passages, schemas, or explicit instructions.
They do not require VLMs to recover the raw data organization before producing structured outputs.
In contrast, our benchmark starts from raw heterogeneous medical resources and evaluates source identification, visual conversion, annotation alignment, metadata organization, and schema-constrained generation in one standardization task.

\section{Task and Benchmark Design}
\subsection{Task Definition}

\begin{figure*}[t]
    \centering
    \includegraphics[width=\linewidth]{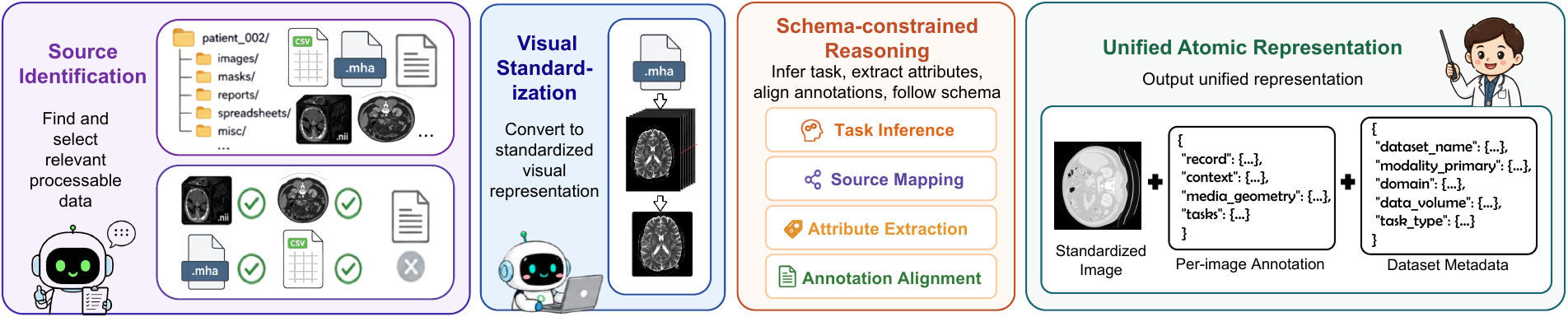}
    \caption{Staged raw medical data standardization workflow, where the agent identifies source evidence, standardizes visual inputs, aligns annotations and metadata, and outputs unified image-text atomic units.}
    \label{fig:reasoning}
\end{figure*}

Raw medical data standardization evaluates whether a VLM can transform raw medical dataset resources into unified, source-grounded representations for downstream use. Given a raw dataset folder and a target schema, the model must identify relevant source files, convert processable visual evidence, align annotations and metadata, and map the recovered information into the required output format. Unlike medical image classification, detection, segmentation, or visual question answering, this task does not assume that the image, label, annotation, and metadata have already been selected and aligned~\cite{sun2016benchmark,litjens2017survey,lin2023medical}. It therefore targets the missing upstream step before conventional medical VLM evaluation.

The required output contains three components:
\begin{itemize}
    \vspace{-2mm}
    \item \textbf{Standardized image:} converted or rendered from the selected raw source(s) and saved in a common visual format, so downstream models can access heterogeneous medical resources through a unified interface.
    \vspace{-2mm}
    \item \textbf{Per-image JSON annotation:} records sample-level semantics and provenance, including dataset identity, modality, anatomical context, task type, spatial attributes, source mapping, and task-specific annotations.
        \vspace{-2mm}
    \item \textbf{Dataset-level JSON annotation:} summarizes global information shared across samples, such as dataset name, modality, data volume, task type, label space, and classes.
        \vspace{-2mm}
\end{itemize}
All annotations must conform to the target schema and remain grounded in evidence from the original raw resources. Figure~\ref{fig:method} summarizes the benchmark pipeline from dataset collection and metadata verification to raw input processing, image-JSON output generation, and eleven-metric evaluation.

\subsection{Benchmark Construction}
Our MDS-Bench is designed to evaluate standardization under raw data conditions rather than on previously curated image and annotation pairs. It contains 1,939 target samples across classification, segmentation, and detection. Each sample requires the VLMs to locate source files or source records, render visual evidence when needed, and map aligned information into the target schema. The resulting benchmark covers common medical vision tasks while retaining the upstream challenges of source selection, visual conversion, annotation alignment, and schema mapping that are often abstracted away in downstream VLM evaluation.

Before selection, we collect approximately 220 candidate datasets from public repositories and official websites and examine their accessibility, task relevance, raw data types, and annotation or metadata availability. Details are provided in Appendix~\ref{app:dataset_selection}. The final 100 datasets are selected to cover diversity in task type, modality, raw format, annotation style, and directory layout. As summarized in Figure~\ref{fig:benchmark_stats}, the benchmark includes \textbf{51} segmentation datasets, \textbf{43} classification datasets, and \textbf{6} detection datasets. The complete dataset catalog is provided in Appendix~\ref{app:dataset_catalog}. It covers major modalities such as CT, MRI, microscopy, pathology, X-ray, ultrasound, ophthalmic imaging, and endoscopy, and supports various raw source formats, including DICOM, TIFF, NIfTI, bitmap images, array files, medical volumes, videos, and specialized formats. This diversity allows us to evaluate whether VLMs can standardize raw medical resources across different storage formats, annotation standards, and directory structures, rather than just adhering to a predefined schema.

Ground-truth construction combines model-assisted extraction with human verification. For model-assisted extraction, we use Gemini 3 Flash because it supports multimodal file understanding and structured output. The model receives the raw dataset folder, sample manifest, target schema, and extraction prompt. It examines the available files and documentation, identifies candidate source files and sample mappings, extracts available metadata and annotations, and drafts standardized JSON records accordingly.

For human verification, six annotators from Computer Science and related disciplines manually verify all 1,939 drafted targets against the original source evidence. All annotators follow the same verification criteria. They check source-file or source-record mapping, image-JSON correspondence, modality and task type, labels, masks, boxes, metadata, and task-specific fields. Unsupported or conflicting values are corrected according to the source evidence, while fields without sufficient evidence are set to null. For quality control, 20\% of the verified cases are independently re-evaluated by another annotator, with coverage across different datasets, tasks, modalities, and file formats. Ambiguous or inconsistent cases are resolved by revisiting the original resources until consensus is reached among annotators.

\subsection{Staged Reasoning Design}
\label{sec:reasoning_design}

Raw medical data standardization cannot be handled as direct schema filling, because the required evidence is scattered across images, volumes, masks, label files, metadata, and directory structures.
We therefore design the prompt as a staged reasoning workflow, as shown in Figure~\ref{fig:reasoning}.
The workflow first guides the agent to locate source evidence, then convert visual content, then reason over annotations and metadata, and finally check whether the generated outputs remain source-grounded and mutually consistent.
This staged design fits the agentic VLM setting, where the model can inspect files, use tools, and run scripts before producing standardized outputs.

The first stage addresses source grounding.
Before generating annotations, the agent must identify which raw files or records support the target sample and separate processable visual sources from auxiliary resources such as masks, label tables, and metadata files.
Once the relevant source evidence is found, the second stage converts visual content into a common image representation.
This conversion is not intended to create a visually plausible image in isolation; instead, it preserves the medical content needed for later annotation reasoning.

The third stage converts grounded evidence into structured annotations. The agent infers task type, modality, dimension, anatomical information, labels, masks, boxes, and provenance based on support from raw files, annotations, metadata, file names, directory structure, or dataset documentation. The final stage checks consistency across the standardized image, the per-image annotation, and the dataset annotation.
Together, these stages connect source identification, visual conversion, annotation alignment, and schema-constrained generation into one reasoning pipeline.

\begin{table}[t]
\centering
\footnotesize
\setlength{\tabcolsep}{3pt}
\renewcommand{\arraystretch}{1.08}
\begin{tabularx}{\columnwidth}{@{}l>{\raggedright\arraybackslash}X@{}}
\toprule
\textbf{Metric} & \textbf{Definition} \\
\midrule
SV  & Basic schema validity, including required blocks, fields, and task-specific branches. \\
SSC & Whether structurally valid JSON is also semantically consistent with the sample. \\
SC  & Correctness of high-level sample meaning, including dataset identity, modality, task type, dimension, class information, and annotation type. \\
IC  & Completeness of expected auxiliary field groups. \\
IV  & Validity of populated auxiliary fields. \\
INR & Degree to which the output avoids redundant or repetitive field filling. \\
CF  & Agreement with verified labels, masks, boxes, source mappings, and task-specific payloads. \\
MSC & Semantic correctness of dataset-level metadata. \\
MSJ & Consistency between dataset-level metadata reliability and sample-level content fidelity. \\
SCJ & Joint success of source selection and content preservation. \\
E2E & Strict full-pipeline success under valid image-JSON pairing, exact source match, schema, semantic, and content constraints. \\
\bottomrule
\end{tabularx}
\caption{Brief definitions of the eleven evaluation metrics used in MDS-Bench, covering structure, semantics, content, metadata, and joint evaluation.}
\label{tab:metric_definitions}
\end{table}

\begin{table*}[t]
  \centering
  \small
  \setlength{\tabcolsep}{3.2pt}
  \renewcommand{\arraystretch}{1.12}

  \resizebox{\textwidth}{!}{%
  \begin{tabular}{l|cc|c|cccc|cc|cc}
    \hline
    \multirow{2}{*}{\textbf{Model}}
    & \multicolumn{2}{c|}{\textbf{Structure}}
    & \multicolumn{1}{c|}{\textbf{Semantic}}
    & \multicolumn{4}{c|}{\textbf{Content}}
    & \multicolumn{2}{c|}{\textbf{Metadata}}
    & \multicolumn{2}{c}{\textbf{Joint}} \\

    & \textbf{SV}
    & \textbf{SSC}
    & \textbf{SC}
    & \textbf{IC}
    & \textbf{IV}
    & \textbf{INR}
    & \textbf{CF}
    & \textbf{MSC}
    & \textbf{MSJ}
    & \textbf{SCJ}
    & \textbf{E2E} \\
    \hline

    Claude Haiku 4.5~\cite{anthropic2025claudehaiku45}
    & 82.2 & 49.8
    & 49.1
    & 47.5 & 41.5 & 42.5 & 40.3
    & 56.8 & 34.1
    & 38.0 & 24.9 \\

    Grok 4.20~\cite{xai2026grok420}
    & 80.0 & 48.8
    & 55.7
    & 49.0 & 45.5 & 46.7 & 41.7
    & 57.8 & 35.6
    & 40.0 & 21.5 \\

    Kimi K2.5~\cite{kimiteam2026kimik25}
    & 83.6 & 51.7
    & 55.1
    & 54.5 & 45.5 & 47.2 & 46.7
    & 63.4 & 40.4
    & 44.8 & 29.4 \\

    GPT-5.2~\cite{openai2025gpt52}
    & 81.6 & 52.8
    & 60.2
    & 50.0 & 37.5 & 47.3 & 47.7
    & \second{74.6} & 43.4
    & 45.4 & 23.6 \\

    GPT-5.2-Codex~\cite{openai2025gpt52codex}
    & 83.7 & 53.3
    & 57.0
    & 55.5 & 49.0 & 48.1 & 47.1
    & 65.6 & 40.6
    & 44.8 & 28.1 \\

    Composer 1.5~\cite{cursor2026composer15}
    & 81.9 & 54.0
    & 59.9
    & 50.5 & 41.3 & 46.8 & 48.5
    & 72.6 & 43.4
    & 46.7 & 30.8 \\

    Claude Sonnet 4.6~\cite{anthropic2026claudesonnet46}
    & \second{85.2} & 56.0
    & 62.5
    & \second{60.0} & \second{56.3} & 50.5 & 49.9
    & 69.0 & 42.7
    & 48.6 & \second{33.8} \\

    Claude Opus 4.6~\cite{anthropic2026claudeopus46}
    & 83.9 & \second{56.2}
    & \best{65.2}
    & 59.5 & 51.8 & \second{51.7} & \second{53.7}
    & 74.4 & \second{48.0}
    & \second{52.3} & 30.5 \\

    Gemini 3 Flash~\cite{googledeepmind2025gemini3flash}
    & \best{88.2} & \best{62.0}
    & \second{63.9}
    & \best{67.5} & \best{62.5} & \best{58.7} & \best{58.7}
    & \best{78.0} & \best{53.7}
    & \best{56.3} & \best{48.6} \\

    \hline
  \end{tabular}%
  }

  \caption{Benchmark performance on the MDS-Bench for nine VLMs with eleven evaluation metrics. All values are percentages (\%) macro-averaged across the 100 datasets. The best result in each column is highlighted in green, and the second-best result is highlighted in light green.}
  \label{tab:main_results}
\end{table*}

Raw medical data standardization is a coupled pipeline rather than a single schema generation task.
A VLM is required to select the correct source, produce a valid structured output, recover the correct medical semantics, preserve task annotations, and keep dataset information consistent with the sample.
If any step fails, the final standardized data may look usable but become unreliable.
Therefore, our evaluation protocol measures both component quality and full-pipeline success.

\subsection{Evaluation Protocol}
\label{sec:evaluation}

We organize the eleven metrics into five capability groups according to the main failure points in this chain. 
\textbf{Structure metrics} include Schema Validity (SV) and Schema Semantic Composite (SSC).
They test whether the generated JSON can serve as a valid standardized record and whether the chosen schema matches the actual sample semantics.
This group is needed because a syntactically valid JSON may still describe the wrong modality, task, or annotation type.

\textbf{Semantic metrics} include Semantic Correctness (SC).
SC evaluates the key fields that determine how the sample should be interpreted, such as dataset identity, modality, task type, data dimension, class information, and annotation type.
Once the interpretation is correct, the next question is whether the sample annotation contains useful and reliable content.
\textbf{Content metrics} include Information Completeness (IC), Information Validity (IV), Information Non-Redundancy (INR), and Content Fidelity (CF).
They measure whether required fields are filled, valid, concise, and faithful to the verified source evidence.

Standardized data also need correct dataset context, not only correct sample annotations.
\textbf{Metadata metrics} include Metadata Semantic Correctness (MSC) and Meta-Sample Joint Score (MSJ).
They evaluate dataset information such as modality, task type, data volume, label availability, and class statistics, and further check whether this dataset information agrees with the generated sample content.
Finally, \textbf{Joint metrics} include Source Content Joint Score (SCJ) and End-to-End Strict Pass (E2E).
They test whether source selection, structure, semantics, and content fidelity succeed together, which gives the strictest view of real standardization success. Table~\ref{tab:metric_definitions} summarizes the definitions of all eleven metrics.

\begin{figure*}[t]
    \centering
    \includegraphics[width=1\linewidth]{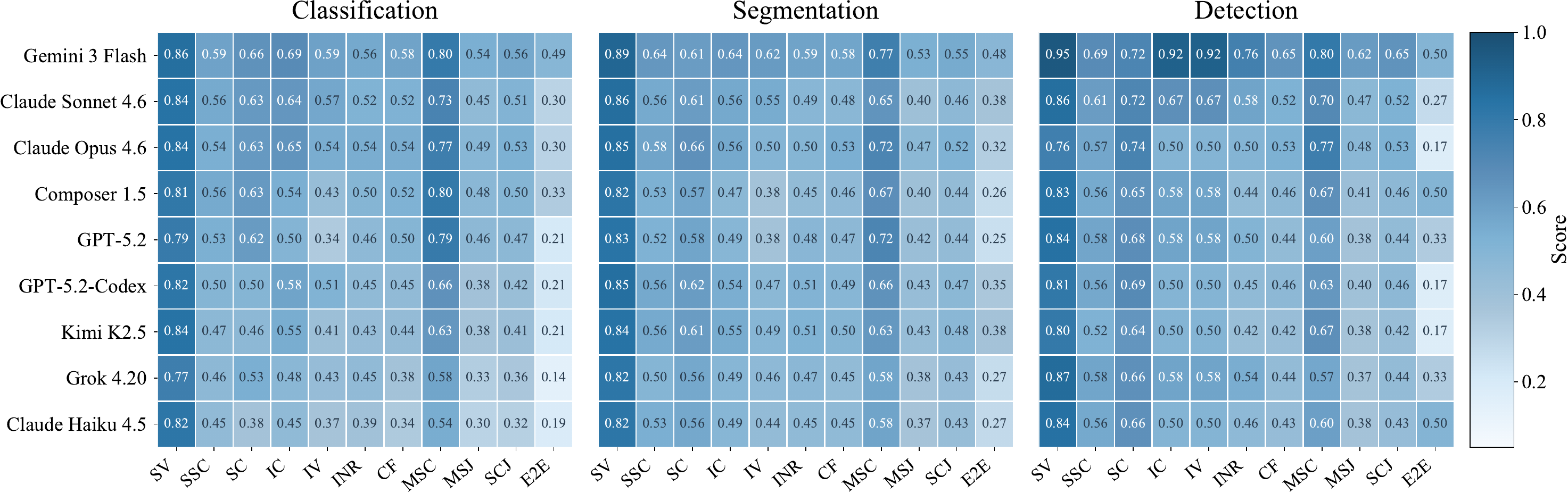}
    \caption{Per-task performance of all evaluated models on the proposed benchmark. Heatmaps report scores on the eleven evaluation metrics for classification, segmentation, and detection subsets, respectively. Rows correspond to models, columns correspond to metrics, and darker colors indicate higher scores.}
    \label{fig:heatmap}
\end{figure*}
Source matching gates field-level evaluation. For a dataset with $N$ target samples, let $A_i^{\mathrm{pred}}$ and $A_i^{\mathrm{gt}}$ denote the normalized predicted and verified source-alias sets, and let $t_i$ indicate whether explicit source-trace evidence is present. The source-matching score is
\begin{equation}
s_i =
\mathbb{I}\left[
A_i^{\mathrm{pred}} \cap A_i^{\mathrm{gt}} \neq \emptyset
\land t_i = 1
\right].
\end{equation}
Thus, $s_i\in\{0,1\}$ equals 1 only when the predicted source matches a verified source alias and has explicit trace evidence. Field metrics SC, IC, IV, INR, and CF are computed only over samples with $s_i=1$, preventing an unrelated source from receiving field-level credit. SV, SSC, SCJ, and E2E are computed over all $N$ samples.

For content fidelity, let $B_i$ denote the applicable expected field blocks for sample $i$, and let $F_{ib}$ denote the verified fields in block $b$. The content-fidelity score is
\begin{equation}
f_i=
\frac{1}{|B_i|}
\sum_{b\in B_i}
\left(
\frac{1}{|F_{ib}|}
\sum_{j\in F_{ib}}\phi_{ibj}
\right),
\end{equation}
where $\phi_{ibj}\in[0,1]$ is the deterministic field-level agreement score.

The remaining per-sample quantities are also deterministically computed: $p_i\in\{0,1\}$ indicates whether the image-JSON pair is valid, $g_i\in[0,1]$ is the fraction of applicable schema checks passed, and $q_i\in[0,1]$ combines semantic-field agreement, required-field presence, and task-specific consistency. The semantic correctness score is
\begin{equation}
\begin{aligned}
q_i ={}& 0.70S_{\mathrm{fields}}(i)
+0.10S_{\mathrm{presence}}(i) \\
&+0.20S_{\mathrm{task}}(i).
\end{aligned}
\end{equation}
where the three terms measure weighted semantic-field agreement, the presence of required task-payload fields, and agreement with the verified task-specific payload, respectively. All per-sample scores are based on deterministic checks; none relies on embedding or cosine similarity. Detailed formulations and underlying checks are provided in Appendix~\ref{app:evaluation_details}.

We define SCJ as
\begin{equation}
\mathrm{SCJ} = \frac{1}{N}\sum_{i=1}^{N} s_i f_i ,
\end{equation}
where $f_i$ is the content fidelity score.
For E2E, let $p_i$ indicate whether the model produces a valid image and JSON pair, $g_i$ denote schema validity, and $q_i$ denote semantic correctness.
The strict pass indicator is
\begin{equation}
\begin{aligned}
e_i = \mathbb{I}\big[
& p_i \land s_i = 1 \land g_i \ge 0.85 \\
& \land q_i \ge 0.5 \land f_i \ge 0.5
\big],
\end{aligned}
\end{equation}
And the final E2E score is
\begin{equation}
\mathrm{E2E} = \frac{1}{N}\sum_{i=1}^{N} e_i .
\end{equation}
We use $g_i \ge 0.85$ rather than perfect schema validity because a minor schema defect should not erase credit from an otherwise correct sample. The default thresholds are selected in preliminary experiments based on downstream usability. We set $q_i\geq0.50$ and $f_i\geq0.50$ as minimum agreement cutoffs, and report threshold sensitivity and ranking stability in Appendix~\ref{app:threshold_analysis}.

All metrics are computed within each dataset and then averaged across the 100 datasets with equal weight.
This prevents large datasets from dominating the benchmark and better reflects robustness across modalities, source formats, annotation styles, and directory layouts. Full metric definitions and details are provided in Appendix~\ref{app:evaluation_details}.

\section{Experiment}

\subsection{Experimental Setup}
We evaluate frontier agentic VLMs on MDS-Bench. Each model receives the raw dataset directory and the target schema, and is required to identify relevant source files or records, convert processable medical data into standardized images, and generate both per-image annotations and a dataset-level annotation. All models are evaluated with the same output format and scoring protocol, and no outputs are manually corrected after generation. We compare Claude Haiku 4.5~\cite{anthropic2025claudehaiku45}, Claude Sonnet 4.6~\cite{anthropic2026claudesonnet46}, Claude Opus 4.6~\cite{anthropic2026claudeopus46}, GPT-5.2~\cite{openai2025gpt52}, GPT-5.2-Codex~\cite{openai2025gpt52codex}, Gemini 3 Flash~\cite{googledeepmind2025gemini3flash}, Grok 4.20~\cite{xai2026grok420}, Kimi K2.5~\cite{kimiteam2026kimik25}, and Composer 1.5~\cite{cursor2026composer15}. For each model, all metrics are first computed within each dataset and then averaged across datasets with equal weight. All models follow the same prompting protocol and agentic setup; implementation details are provided in Appendix~\ref{app:agentic_setup} and our public code repository.

\begin{table*}[t]
  \centering
  \small
  \setlength{\tabcolsep}{3.2pt}
  \renewcommand{\arraystretch}{1.12}

  \resizebox{\textwidth}{!}{%
  \begin{tabular}{l|cc|cc|cc|cc|cc}
    \hline
    \multirow{2}{*}{\textbf{Model}}
    & \multicolumn{2}{c|}{\textbf{Structure}}
    & \multicolumn{2}{c|}{\textbf{Semantic}}
    & \multicolumn{2}{c|}{\textbf{Content}}
    & \multicolumn{2}{c|}{\textbf{Metadata}}
    & \multicolumn{2}{c}{\textbf{Joint}} \\

    & \textbf{Score (\%)} & \textbf{\# Fail}
    & \textbf{Score (\%)} & \textbf{\# Fail}
    & \textbf{Score (\%)} & \textbf{\# Fail}
    & \textbf{Score (\%)} & \textbf{\# Fail}
    & \textbf{Score (\%)} & \textbf{\# Fail} \\
    \hline

    Claude Haiku 4.5~\cite{anthropic2025claudehaiku45}
    & 66.0 & 10
    & 49.1 & 41
    & 42.9 & 58
    & 45.4 & 57
    & 31.5 & 75 \\

    Grok 4.20~\cite{xai2026grok420}
    & 64.4 & 13
    & 55.7 & 31
    & 45.7 & 64
    & 46.7 & 63
    & 30.7 & 78 \\

    Kimi K2.5~\cite{kimiteam2026kimik25}
    & 67.7 & 9
    & 55.1 & 35
    & 48.5 & 54
    & 51.9 & 52
    & 37.1 & 69 \\

    GPT-5.2~\cite{openai2025gpt52}
    & 67.2 & 4
    & 60.2 & 25
    & 45.6 & 67
    & 59.0 & \second{34}
    & 34.5 & 76 \\

    GPT-5.2-Codex~\cite{openai2025gpt52codex}
    & 68.5 & 5
    & 57.0 & 27
    & 49.9 & 49
    & 53.1 & 44
    & 36.5 & 72 \\

    Composer 1.5~\cite{cursor2026composer15}
    & 67.9 & 6
    & 59.9 & 25
    & 46.8 & 53
    & 58.0 & 36
    & 38.7 & 70 \\

    Claude Sonnet 4.6~\cite{anthropic2026claudesonnet46}
    & \second{70.6} & \best{1}
    & 62.5 & \second{23}
    & \second{54.2} & \second{45}
    & 55.8 & 43
    & 41.2 & \second{65} \\

    Claude Opus 4.6~\cite{anthropic2026claudeopus46}
    & 70.0 & \second{3}
    & \best{65.2} & \best{20}
    & \second{54.2} & \second{45}
    & \second{61.2} & \second{34}
    & \second{41.4} & \second{65} \\

    Gemini 3 Flash~\cite{googledeepmind2025gemini3flash}
    & \best{75.1} & 4
    & \second{63.9} & \second{23}
    & \best{61.8} & \best{39}
    & \best{65.9} & \best{29}
    & \best{52.4} & \best{51} \\

    \hline
  \end{tabular}%
  }

  \caption{Error attribution across five capability categories. For each category, Score reports the mean category score, and \# Fail reports the number of datasets whose category score is below 50\%. Higher Score and lower \# Fail indicate better performance.}
  \label{tab:error_attribution}
\end{table*}

\begin{figure}[t]
    \centering
    \includegraphics[width=1\linewidth]{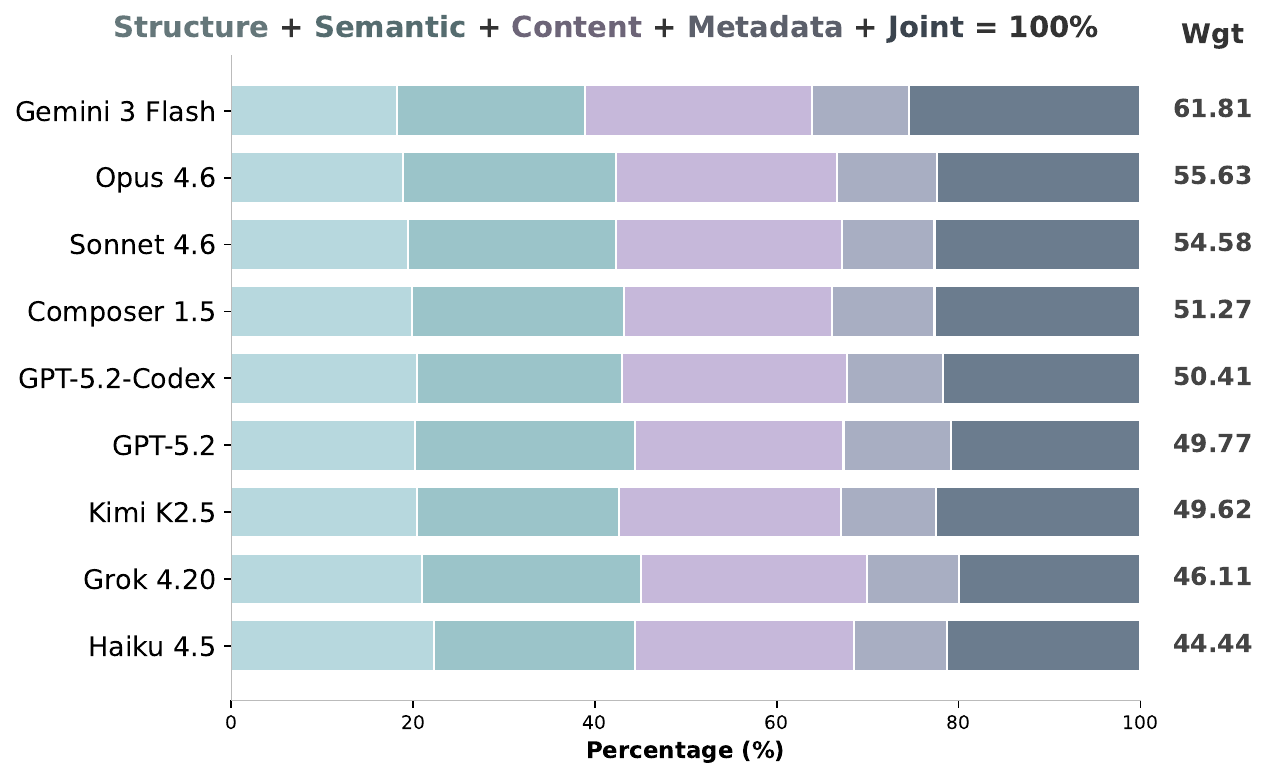}
\caption{Weighted capability composition on MDS-Bench. Each stacked bar decomposes a model's normalized performance into Structure, Semantic, Content, Metadata, and Joint categories using weights 0.15, 0.20, 0.25, 0.10, and 0.30. The rightmost value reports the overall weighted score (Wgt).}
    \label{fig:weighted_capability}
\end{figure}

\vspace{-1mm}

\subsection{Overall Results}
\label{sec:overall_results}

\textbf{Models can often satisfy the output schema, but schema validity does not imply correct standardization.}
As shown in Table~\ref{tab:main_results}, most models obtain relatively high SV, ranging from 80.0\% to 88.2\%.
However, SSC is much lower than expected, ranging from 48.8\% to 62.0\%.
This gap shows that models can generate JSON files with the required fields, while still making mistakes in task branch, modality, dataset identity, or annotation semantics.
Thus, schema following is only a basic requirement for raw medical data standardization, not a reliable indicator of task success.

\textbf{Joint metrics show that current models still fail to complete the full standardization pipeline reliably.}
SCJ ranges from 38.0\% to 56.3\%, and E2E ranges from 21.5\% to 48.6\%.
These scores are much lower than structure metrics because they require several conditions to hold together: a valid image-text pair, correct source selection, sufficient schema quality, semantic correctness, and content fidelity.
The low joint scores indicate that models can often solve individual parts of the task, but errors accumulate when the complete raw-data-to-standardized-output pipeline is evaluated.

\textbf{Gemini 3 Flash is the strongest overall model, but even the best model remains far from reliable end-to-end standardization.}
Gemini 3 Flash obtains the best score on most metrics, including SV, SSC, IC, IV, INR, CF, MSC, MSJ, SCJ, and E2E.
Claude Opus 4.6 achieves the highest SC, suggesting stronger recovery of key task semantics.
Nevertheless, Gemini 3 Flash reaches only 48.6\% E2E, meaning that more than half of the target samples still fail at least one required condition.
This result highlights the gap between strong component-level performance and reliable full-pipeline standardization.

\textbf{The same structure-joint gap appears across task subsets, while model rankings vary under different data conditions.}
Figure~\ref{fig:heatmap} shows that schema-level metrics remain higher than joint metrics across classification, segmentation, and detection, with E2E remaining difficult in all three task types.
Additional cross-group results in Appendix~\ref{app:cross_group} show that Gemini 3 Flash is the most robust model overall, while Claude Opus 4.6 performs best on Microscopy / Pathology datasets.

\textbf{Weighted capability composition further shows that overall performance is driven by content and joint reliability.}
Figure~\ref{fig:weighted_capability} summarizes model performance through a capability-weighted decomposition. Since Joint and Content receive the largest weights, models with stronger full-pipeline consistency and content fidelity obtain higher overall weighted scores. Gemini 3 Flash ranks first, indicating that its advantage comes not only from schema following but also from stronger content recovery and joint reliability. In contrast, models with reasonable Structure or Semantic contributions can still rank lower when their Joint component is weak. This pattern reinforces our main finding that raw medical data standardization is primarily bottlenecked by source-grounded content preservation and end-to-end consistency.

\begin{figure*}[ht]
    \centering
    \includegraphics[width=\textwidth]{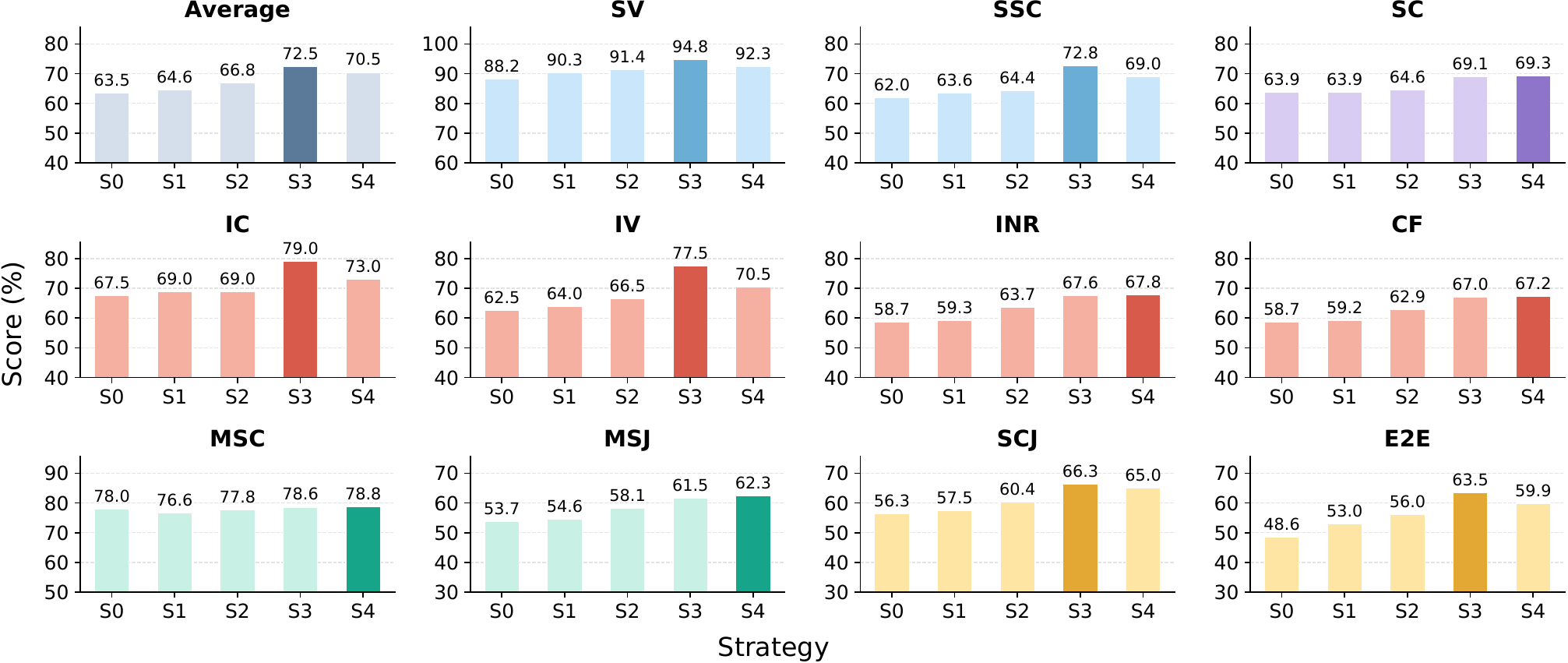}
    \caption{Inference strategy ablation on MDS-Bench using Gemini 3 Flash as the base model. Strategies S0-S4 compare direct generation with increasingly validation-guided refinement and selection procedures. The first panel reports the average score across all eleven metrics, while the remaining panels show metric-specific changes in structure, semantic, content, metadata, and joint performance.}
    \label{fig:strategy_ablation}
\end{figure*}

\subsection{Error Attribution Analysis}

\textbf{Dataset-level failure rates show that errors are concentrated in Content and Joint, while Structure failures are rare.}
Table~\ref{tab:error_attribution} groups the eleven metrics into five categories and reports both the mean category score and the percentage of datasets whose category score falls below 0.5.
Structure has the lowest failure rate for all models, ranging from 1\% to 13\%, which means that most datasets do not fail because the model breaks the output format.
In contrast, Content and Joint have much higher failure rates.
This pattern indicates that the primary failures arise when models are required to maintain labels, masks, boxes, source records, and task-specific annotation payloads, as opposed to simply adhering to the schema.
This distinction is also visible in individual outputs, which reveal four representative failure modes:
\begin{itemize}
\setlength{\itemsep}{0pt}
\setlength{\parsep}{0pt}
\setlength{\parskip}{0pt}
\raggedright

\item \textbf{Source grounding:} Claude Sonnet 4.6 selects an incorrect \texttt{.mat} source for \texttt{human-protein-atlas-cells-dataset} instead of the \texttt{.parquet} record.

\item \textbf{Semantic confusion:} GPT-5.2-Codex treats a segmentation sample in \texttt{melanoma-train} as classification.

\item \textbf{Content omission:} Gemini 3 Flash omits several required fields for a sample from \texttt{PBC\_Brain\_MRI\_5005236} despite correctly identifying the task and label.

\item \textbf{Unsupported filling:} Composer 1.5 fills the 2D, 3D, and video branches simultaneously for \texttt{NYU\_POAG\_RetinalOCT\_Glaucoma}.

\end{itemize}

\textbf{The category breakdown also shows that model strengths do not transfer uniformly across levels of standardization.}
Gemini 3 Flash has the strongest overall error profile, but other models show different strengths and weaknesses.
Claude Opus 4.6 achieves the best Semantic score, yet its Joint failure rate remains 65\%.
GPT-5.2 and Composer 1.5 obtain relatively strong Metadata scores, but both remain weak in Content and Joint.
This suggests that models may recover dataset-level context or key task semantics while still failing to preserve reliable sample-level content.
The key takeaway from this analysis is that enhancing Structure or Metadata alone is inadequate; effective standardization of raw medical data necessitates improved Content recovery and greater Joint reliability across datasets.

\subsection{Inference Strategy Ablation}
\label{sec:inference_ablation}

\textbf{Validation-based inference improves overall standardization quality.} 
Figure~\ref{fig:strategy_ablation} compares five strategies using Gemini 3 Flash. S0 directly generates the standardized output, S1 adds self-refinement, S2 uses verification-guided refinement, S3 selects the best complete candidate, and S4 performs validator-guided field-wise selection. Detailed definitions and validation procedures are provided in Appendix~\ref{app:inference_strategies}. Relative to S0, S1 and S2 increase E2E from 48.6\% to 53.0\% and 56.0\%, while SCJ increases from 56.3\% to 57.5\% and 60.4\%. This demonstrates that additional verification can minimize certain source and content-related errors without altering the base model.

\textbf{S3 is most effective because complete-candidate selection yields the largest gains on joint metrics.}
S3 achieves the best overall results, with 66.3\% SCJ and 63.5\% E2E, improving E2E by 14.9 points over S0.
It also obtains the highest scores on SV, SSC, IC, and IV.
These gains suggest that selecting the best full output is especially useful for raw data standardization, where source correctness, schema quality, semantic correctness, and content fidelity must hold at the same time.

\textbf{Field-wise selection improves local quality but not end-to-end reliability.} 
S4 obtains the best scores on SC, INR, CF, MSC, and MSJ, indicating stronger local semantic, content, and metadata fields.
However, its SCJ and E2E are lower than S3, at 65.0\% and 59.9\%.
This suggests that combining locally better fields does not necessarily produce the most coherent standardized sample.
Overall, the ablation results indicate that selecting a comprehensive candidate is more effective than independently patching fields.

\vspace{-1mm}

\section{Conclusion}

We introduced raw medical data standardization as an upstream task for medical AI and built MDS-Bench to evaluate VLM agents on heterogeneous raw medical resources. Experiments show that current models can often satisfy schema-level requirements, but still struggle with Content and Joint, leading to low E2E performance even for Gemini 3 Flash. Validation-based inference improves results, especially through complete-candidate selection, but does not eliminate the core bottleneck. These findings show that reliable medical AI requires not only structured output generation, but also faithful content recovery and source-aligned standardization from raw data.

\clearpage

\section*{Limitations} This work has several limitations. The benchmark is built from publicly available medical imaging datasets and may not fully reflect real clinical data environments with private PACS systems, electronic health records, access-control constraints, or incomplete metadata. It focuses on classification, detection, and segmentation, leaving other medical AI settings such as report generation, longitudinal modeling, and treatment-oriented decision support for future work. The evaluation protocol uses schema-based and source-grounded checks, which may miss subtle medical interpretation errors or penalize alternative valid outputs. Ground truth construction relies on large-model-assisted extraction followed by human verification, so residual annotation errors may remain. Although we use independent re-evaluation and consensus-based resolution for quality control, the pre-resolution annotations are not retained, preventing retrospective computation of a formal inter-annotator agreement score. In addition, inference parameters such as temperature, top-p, and maximum token limits are controlled by the providers of some commercial closed-source models and are not publicly exposed, so we cannot manually configure or consistently report these settings across models. The benchmark is intended for research evaluation only, and model-generated standardized medical data should not be used for clinical decision making without expert verification.

\bibliography{custom}

@misc{anthropic2025claudehaiku45,
  title  = {{Claude Haiku 4.5} System Card},
  author = {{Anthropic}},
  year   = {2025},
  month  = oct,
  url    = {https://www-cdn.anthropic.com/7aad69bf12627d42234e01ee7c36305dc2f6a970.pdf}
}

@misc{anthropic2026claudesonnet46,
  title  = {{Claude Sonnet 4.6} System Card},
  author = {{Anthropic}},
  year   = {2026},
  month  = feb,
  url    = {https://www-cdn.anthropic.com/78073f739564e986ff3e28522761a7a0b4484f84.pdf}
}

@misc{anthropic2026claudeopus46,
  title  = {{Claude Opus 4.6} System Card},
  author = {{Anthropic}},
  year   = {2026},
  month  = feb,
  url    = {https://www-cdn.anthropic.com/0dd865075ad3132672ee0ab40b05a53f14cf5288.pdf}
}

@misc{openai2025gpt52,
  title  = {Update to {GPT-5} System Card: {GPT-5.2}},
  author = {{OpenAI}},
  year   = {2025},
  month  = dec,
  url    = {https://openai.com/index/gpt-5-system-card-update-gpt-5-2/}
}

@misc{openai2025gpt52codex,
  title  = {Addendum to {GPT-5.2} System Card: {GPT-5.2-Codex}},
  author = {{OpenAI}},
  year   = {2025},
  url    = {https://openai.com/index/gpt-5-2-codex-system-card/}
}

@article{deng2026project,
  title={{Project Imaging-X}: A survey of 1000+ open-access medical imaging datasets for foundation model development},
  author={Deng, Zhongying and Tang, Cheng and Huang, Ziyan and Lin, Jiashi and Chen, Ying and Ning, Junzhi and Ma, Chenglong and Liu, Jiyao and Li, Wei and Zhu, Yinghao and others},
  journal={arXiv preprint arXiv:2603.27460},
  year={2026}
}

@misc{googledeepmind2025gemini3flash,
  title  = {{Gemini 3 Flash} Model Card},
  author = {{Google DeepMind}},
  year   = {2025},
  month  = dec,
  url    = {https://storage.googleapis.com/deepmind-media/Model-Cards/Gemini-3-Flash-Model-Card.pdf}
}

@misc{xai2026grok420,
  title  = {{Grok 4.20} Model Documentation},
  author = {{xAI}},
  year   = {2026},
  url    = {https://docs.x.ai/developers/models/grok-4.20}
}

@misc{cursor2026composer15,
  title  = {Introducing {Composer 1.5}},
  author = {{Cursor}},
  year   = {2026},
  month  = feb,
  url    = {https://cursor.com/blog/composer-1-5}
}

@article{kimiteam2026kimik25,
  title   = {{Kimi K2.5}: Visual Agentic Intelligence},
  author  = {{Kimi Team} and Bai, Tongtong and Bai, Yifan and Bao, Yiping and Cai, S. H. and Cao, Yuan and Charles, Y. and Che, H. S. and Chen, Cheng and Chen, Guanduo and others},
  journal = {arXiv preprint arXiv:2602.02276},
  year    = {2026},
  url     = {https://arxiv.org/abs/2602.02276}
}

@article{singhal2023large,
  title={Large language models encode clinical knowledge},
  author={Singhal, Karan and Azizi, Shekoofeh and Tu, Tao and Mahdavi, S Sara and Wei, Jason and Chung, Hyung Won and Scales, Nathan and Tanwani, Ajay and Cole-Lewis, Heather and Pfohl, Stephen and others},
  journal={Nature},
  volume={620},
  number={7972},
  pages={172--180},
  year={2023},
  publisher={Nature Publishing Group UK London}
}

@inproceedings{sun2016benchmark,
  title={A benchmark for automatic visual classification of clinical skin disease images},
  author={Sun, Xiaoxiao and Yang, Jufeng and Sun, Ming and Wang, Kai},
  booktitle={European conference on computer vision},
  pages={206--222},
  year={2016},
  organization={Springer}
}

@inproceedings{d2026mmbu,
  title={{MMBU}: A Massive Multi-modal Biomedical Understanding Benchmark to Probe the Perception Capabilities of Vision-Language Models},
  author={D'Cunha, Ryan and Lozano, Alejandro and Sun, Xiaoxiao and Jarquin, Daniel Vela and Sun, Min Woo and Aklilu, Josiah and Burgess, James and Zhang, Yuhui and Nayebi, Ryan and Avila Robayo, Paola and others},
  booktitle={European Conference on Computer Vision (ECCV)},
  year={2026}
}

@inproceedings{nentidis2025overview,
  title={Overview of {BioASQ} 2025: The thirteenth {BioASQ} challenge on large-scale biomedical semantic indexing and question answering},
  author={Nentidis, Anastasios and Katsimpras, Georgios and Krithara, Anastasia and Krallinger, Martin and Rodr{\'\i}guez-Ortega, Miguel and Rodriguez-L{\'o}pez, Eduard and Loukachevitch, Natalia and Sakhovskiy, Andrey and Tutubalina, Elena and Dimitriadis, Dimitris and others},
  booktitle={International Conference of the Cross-Language Evaluation Forum for European Languages},
  pages={173--198},
  year={2025},
  organization={Springer}
}

@article{zhang2023pmc,
  title= {{PMC-VQA}: Visual instruction tuning for medical visual question answering},
  author={Zhang, Xiaoman and Wu, Chaoyi and Zhao, Ziheng and Lin, Weixiong and Zhang, Ya and Wang, Yanfeng and Xie, Weidi},
  journal={arXiv preprint arXiv:2305.10415},
  year={2023}
}

@article{chen2024gmai,
  title={{GMAI-MMBench}: A comprehensive multimodal evaluation benchmark towards general medical ai},
  author={Chen, Pengcheng and Ye, Jin and Wang, Guoan and Li, Yanjun and Deng, Zhongying and Li, Wei and Li, Tianbin and Duan, Haodong and Huang, Ziyan and Su, Yanzhou and others},
  journal={Advances in Neural Information Processing Systems},
  volume={37},
  pages={94327--94427},
  year={2024}
}

@inproceedings{yue2024mmmu,
  title={{MMMU}: A massive multi-discipline multimodal understanding and reasoning benchmark for expert agi},
  author={Yue, Xiang and Ni, Yuansheng and Zhang, Kai and Zheng, Tianyu and Liu, Ruoqi and Zhang, Ge and Stevens, Samuel and Jiang, Dongfu and Ren, Weiming and Sun, Yuxuan and others},
  booktitle={Proceedings of the IEEE/CVF conference on computer vision and pattern recognition},
  pages={9556--9567},
  year={2024}
}

@inproceedings{tang2024medagents,
  title={{MedAgents}: Large language models as collaborators for zero-shot medical reasoning},
  author={Tang, Xiangru and Zou, Anni and Zhang, Zhuosheng and Li, Ziming and Zhao, Yilun and Zhang, Xingyao and Cohan, Arman and Gerstein, Mark},
  booktitle={Findings of the Association for Computational Linguistics: ACL 2024},
  pages={599--621},
  year={2024}
}

@article{karargyris2023federated,
  title={Federated benchmarking of medical artificial intelligence with {MedPerf}},
  author={Karargyris, Alexandros and Umeton, Renato and Sheller, Micah J and Aristizabal, Alejandro and George, Johnu and Wuest, Anna and Pati, Sarthak and Kassem, Hasan and Zenk, Maximilian and Baid, Ujjwal and others},
  journal={Nature machine intelligence},
  volume={5},
  number={7},
  pages={799--810},
  year={2023},
  publisher={Nature Publishing Group UK London}
}

@article{yang2025structeval,
  title={{StructEval}: Benchmarking {LLMs}' Capabilities to Generate Structural Outputs},
  author={Yang, Jialin and Jiang, Dongfu and He, Lipeng and Siu, Sherman and Zhang, Yuxuan and Liao, Disen and Li, Zhuofeng and Zeng, Huaye and Jia, Yiming and Wang, Haozhe and others},
  journal={arXiv preprint arXiv:2505.20139},
  year={2025}
}

@article{geng2025jsonschemabench,
  title={{JSONSchemaBench}: A rigorous benchmark of structured outputs for language models},
  author={Geng, Saibo and Cooper, Hudson and Moskal, Micha{\l} and Jenkins, Samuel and Berman, Julian and Ranchin, Nathan and West, Robert and Horvitz, Eric and Nori, Harsha},
  journal={arXiv preprint arXiv:2501.10868},
  year={2025}
}

@article{roberts2024image2struct,
  title={{Image2struct}: Benchmarking structure extraction for vision-language models},
  author={Roberts, Josselin S and Lee, Tony and Wong, Chi H and Yasunaga, Michihiro and Mai, Yifan and Liang, Percy},
  journal={Advances in Neural Information Processing Systems},
  volume={37},
  pages={115058--115097},
  year={2024}
}

@article{feng2025so,
  title={{SO-Bench}: A Structural Output Evaluation of Multimodal {LLMs}},
  author={Feng, Di and Ma, Kaixin and Nan, Feng and Chen, Haofeng and Zhai, Bohan and Griffiths, David and Gao, Mingfei and Gan, Zhe and Verma, Eshan and Yang, Yinfei and others},
  journal={arXiv preprint arXiv:2511.21750},
  year={2025}
}

@article{litjens2017survey,
  title={A survey on deep learning in medical image analysis},
  author={Litjens, Geert and Kooi, Thijs and Bejnordi, Babak Ehteshami and Setio, Arnaud Arindra Adiyoso and Ciompi, Francesco and Ghafoorian, Mohsen and Van Der Laak, Jeroen Awm and Van Ginneken, Bram and S{\'a}nchez, Clara I},
  journal={Medical image analysis},
  volume={42},
  pages={60--88},
  year={2017},
  publisher={Elsevier}
}

@article{li2023llava,
  title={{LLaVA-Med}: Training a large language-and-vision assistant for biomedicine in one day},
  author={Li, Chunyuan and Wong, Cliff and Zhang, Sheng and Usuyama, Naoto and Liu, Haotian and Yang, Jianwei and Naumann, Tristan and Poon, Hoifung and Gao, Jianfeng},
  journal={Advances in Neural Information Processing Systems},
  volume={36},
  pages={28541--28564},
  year={2023}
}

@article{wasserthal2023totalsegmentator,
  title={{TotalSegmentator}: robust segmentation of 104 anatomic structures in {CT} images},
  author={Wasserthal, Jakob and Breit, Hanns-Christian and Meyer, Manfred T and Pradella, Maurice and Hinck, Daniel and Sauter, Alexander W and Heye, Tobias and Boll, Daniel T and Cyriac, Joshy and Yang, Shan and others},
  journal={Radiology: Artificial Intelligence},
  volume={5},
  number={5},
  pages={e230024},
  year={2023},
  publisher={Radiological Society of North America}
}

@article{yao2022react,
  title={{ReAct}: Synergizing reasoning and acting in language models},
  author={Yao, Shunyu and Zhao, Jeffrey and Yu, Dian and Du, Nan and Shafran, Izhak and Narasimhan, Karthik and Cao, Yuan},
  journal={arXiv preprint arXiv:2210.03629},
  year={2022}
}

@article{lin2023medical,
  title={Medical visual question answering: A survey},
  author={Lin, Zhihong and Zhang, Donghao and Tao, Qingyi and Shi, Danli and Haffari, Gholamreza and Wu, Qi and He, Mingguang and Ge, Zongyuan},
  journal={Artificial Intelligence in Medicine},
  volume={143},
  pages={102611},
  year={2023},
  publisher={Elsevier}
}

@article{wilkinson2016fair,
  title={The {FAIR} Guiding Principles for scientific data management and stewardship},
  author={Wilkinson, Mark D and Dumontier, Michel and Aalbersberg, IJsbrand Jan and Appleton, Gabrielle and Axton, Myles and Baak, Arie and Blomberg, Niklas and Boiten, Jan-Willem and da Silva Santos, Luiz Bonino and Bourne, Philip E and others},
  journal={Scientific data},
  volume={3},
  number={1},
  pages={1--9},
  year={2016},
  publisher={Nature Publishing Group}
}

@article{hosseinzadeh2025data,
  title={Data quality assessment in healthcare, dimensions, methods and tools: a systematic review},
  author={Hosseinzadeh, Elham and Afkanpour, Marziyeh and Momeni, Mehri and Tabesh, Hamed},
  journal={BMC Medical Informatics and Decision Making},
  volume={25},
  number={1},
  pages={296},
  year={2025},
  publisher={Springer}
}

@article{pezoulas2019medical,
  title={Medical data quality assessment: On the development of an automated framework for medical data curation},
  author={Pezoulas, Vasileios C and Kourou, Konstantina D and Kalatzis, Fanis and Exarchos, Themis P and Venetsanopoulou, Aliki and Zampeli, Evi and Gandolfo, Saviana and Skopouli, Fotini and De Vita, Salvatore and Tzioufas, Athanasios G and others},
  journal={Computers in biology and medicine},
  volume={107},
  pages={270--283},
  year={2019},
  publisher={Elsevier}
}

@article{barret2025etl,
  title={I-ETL: an interoperability-aware health (meta) data pipeline to enable federated analyses},
  author={Barret, Nelly and Bernasconi, Anna and Bikbov, Boris and Pinoli, Pietro},
  journal={BMC Medical Informatics and Decision Making},
  volume={25},
  number={1},
  pages={375},
  year={2025},
  publisher={Springer}
}

@article{zhang2022heterogeneous,
  title={A heterogeneous multi-modal medical data fusion framework supporting hybrid data exploration},
  author={Zhang, Yong and Sheng, Ming and Liu, Xingyue and Wang, Ruoyu and Lin, Weihang and Ren, Peng and Wang, Xia and Zhao, Enlai and Song, Wenchao},
  journal={Health Information Science and Systems},
  volume={10},
  number={1},
  pages={22},
  year={2022},
  publisher={Springer}
}

@inproceedings{liu2025review,
  title={A review of multimodal medical data fusion techniques for personalized medicine},
  author={Liu, Congcong and Ye, Fangfang},
  booktitle={Proceedings of the 4th International Conference on Biomedical and Intelligent Systems},
  pages={338--347},
  year={2025}
}

@inproceedings{akella2025data,
  title={Data wrangling task automation using code-generating language models},
  author={Akella, Ashlesha and Narayanam, Krishnasuri},
  booktitle={Proceedings of the AAAI Conference on Artificial Intelligence},
  volume={39},
  number={28},
  pages={29616--29618},
  year={2025}
}

@inproceedings{li2025autodcworkflow,
  title={{AutoDCWorkflow}: {LLM-based} Data Cleaning Workflow Auto-Generation and Benchmark.},
  author={Li, Lan and Fang, Liri and Lud{\"a}scher, Bertram and Torvik, Vetle I},
  booktitle={EMNLP (Findings)},
  pages={7766--7780},
  year={2025}
}

@article{jiang2025medagentbench,
  title={{MedAgentBench}: a virtual {EHR} environment to benchmark medical {LLM} agents},
  author={Jiang, Yixing and Black, Kameron C and Geng, Gloria and Park, Danny and Zou, James and Ng, Andrew Y and Chen, Jonathan H},
  journal={{NEJM AI}},
  volume={2},
  number={9},
  pages={AIdbp2500144},
  year={2025},
  publisher={Massachusetts Medical Society}
}
\newpage

\appendix

\section{Dataset Collection and Full Dataset Catalog}
\label{app:collection_and_dataset_catalog}

\subsection{Dataset Collection and Selection}
\label{app:dataset_selection}
Before selection, we collect approximately 220 candidate datasets from public platforms such as Hugging Face, Kaggle, Zenodo, and TCIA, as well as official dataset and challenge websites. We examine their official descriptions and download pages to determine data accessibility, task relevance, raw data types, and the availability of annotations or metadata. Some datasets already provide standard image files and require no conversion, whereas others provide medical-image formats that require further processing. Datasets requiring access approval are considered only when access is available, and eligible datasets are downloaded from their official sources. For each candidate, we inspect the raw files, annotations, metadata, and directory structures. We exclude datasets that remain inaccessible because of access restrictions, contain no usable annotation files after download, have corrupted or incomplete resources, or lack reliable mappings between samples and annotations. Duplicate or highly overlapping versions of the same data are also removed.

\subsection{Full Dataset Catalog} \label{app:dataset_catalog}

Tables~\ref{tab:dataset_catalog_part1}-\ref{tab:dataset_catalog_part3} list the complete set of datasets used in our benchmark. Some Zenodo-derived datasets are listed using their record-ID-based raw folder names, such as entries ending with \texttt{\_folder}. 
These names are retained to preserve direct traceability to the original downloaded raw folders.

\begin{table*}[t]
\centering
\footnotesize
\setlength{\tabcolsep}{3.0pt}
\renewcommand{\arraystretch}{0.98}
\begin{tabular}{@{}p{0.54\textwidth}p{0.19\textwidth}p{0.14\textwidth}p{0.08\textwidth}@{}}
\toprule
\textbf{Dataset name} & \textbf{Modality} & \textbf{Task type} & \textbf{Format} \\
\midrule
Micro\_\allowbreak{}Ultrasound\_\allowbreak{}Prostate\_\allowbreak{}Segmentation\_\allowbreak{}Dataset & Micro-Ultrasound & segmentation & nii \\
BONBID-HIE & MRI & segmentation & mha \\
4939348\_\allowbreak{}Elastographic\_\allowbreak{}Map\_\allowbreak{}Simulation & Ultrasound & classification & mat \\
4975777\_\allowbreak{}folder & Microscopy & classification & czi \\
RetinalAVM\_\allowbreak{}CaseStudy\_\allowbreak{}5002837 & Ophthalmic & classification & tif \\
PBC\_\allowbreak{}Brain\_\allowbreak{}MRI\_\allowbreak{}5005236 & MRI & classification & nii.gz \\
Organoid\_\allowbreak{}MRI\_\allowbreak{}Longitudinal & MRI & classification & dcm \\
Placenta\_\allowbreak{}HSI & Hyperspectral & segmentation & tif \\
15680730\_\allowbreak{}folder & CT & classification & mha \\
Granger2018\_\allowbreak{}AOSLO\_\allowbreak{}RPE\_\allowbreak{}PR\_\allowbreak{}ROIs & Adaptive optics & segmentation & tif \\
SLIMfast4C\_\allowbreak{}HeLa\_\allowbreak{}meGFP\_\allowbreak{}MBP\_\allowbreak{}TMD\_\allowbreak{}Zenodo5712332 & Microscopy & classification & tif \\
Zenodo\_\allowbreak{}5903190\_\allowbreak{}Multiplex\_\allowbreak{}Tissue\_\allowbreak{}Imaging & Microscopy & segmentation & tiff \\
6355622\_\allowbreak{}folder & NA & segmentation & mat \\
NYU\_\allowbreak{}POAG\_\allowbreak{}RetinalOCT\_\allowbreak{}Glaucoma & OCT & classification & mha \\
ICIAR2018\_\allowbreak{}BACH\_\allowbreak{}BreastHistology & Microscopy & classification & tif \\
BACH\_\allowbreak{}ICAR\_\allowbreak{}2018 & Microscopy & classification & tif \\
Brain\_\allowbreak{}MRI\_\allowbreak{}segmentation & MRI & segmentation & tif \\
breast-cancer-microwave & Microwave imaging & classification & npy \\
Brain\_\allowbreak{}Tumor\_\allowbreak{}Dataset & MRI & classification & tif \\
capsule\_\allowbreak{}endoscopy\_\allowbreak{}dataset\_\allowbreak{}kauhc2025 & Capsule endoscopy & classification & bmp \\
Cervical\_\allowbreak{}Cancer\_\allowbreak{}largest\_\allowbreak{}dataset & Microscopy & classification & bmp \\
chest-xray-unidatapro & X-ray & detection & dcm \\
colorectal\_\allowbreak{}histology\_\allowbreak{}mnist2019 & Histopathology & classification & tif \\
ct-of-the-spine-scoliosis & CT & classification & dcm \\
edd2020\_\allowbreak{}endoscopy\_\allowbreak{}detection\_\allowbreak{}and\_\allowbreak{}segmentation2025 & Endoscopy & segmentation & tif \\
expansion-microscopy & Microscopy & segmentation & tif \\
fazekas-mri & MRI & classification & dcm \\
fomo-mri & MRI & classification & nii.gz \\
fundus-image-segmentation & Fundus & segmentation & tif \\
gastrointestinal\_\allowbreak{}bleeding\_\allowbreak{}images\_\allowbreak{}dataset2024 & Endoscopy & classification & bmp \\
human-protein-atlas-cells-dataset & Microscopy & segmentation & parquet \\
isic-skin-cancer-train-npz & Dermoscopy & segmentation & npz \\
kits23-kidney-cancer & CT & segmentation & nii.gz \\
leukemia-cancer-dataset & Microscopy & classification & bmp \\
\bottomrule
\end{tabular}
\caption{\label{tab:dataset_catalog_part1}
Complete dataset catalog used in the proposed benchmark (Part 1 of 3).
}
\end{table*}

\begin{table*}[t]
\centering
\footnotesize
\setlength{\tabcolsep}{3.0pt}
\renewcommand{\arraystretch}{0.98}
\begin{tabular}{@{}p{0.54\textwidth}p{0.19\textwidth}p{0.14\textwidth}p{0.08\textwidth}@{}}
\toprule
\textbf{Dataset name} & \textbf{Modality} & \textbf{Task type} & \textbf{Format} \\
\midrule
medical-7-2020 & MRI & classification & mat \\
melanoma-train & Dermoscopy & segmentation & bmp \\
microscopy-demo & Microscopy & classification & tif \\
mri-rd & MRI & classification & dcm \\
npz-histopathology-dataset & Microscopy & classification & npz \\
pancreas-mri & MRI & classification & dcm \\
tcga-lgg-mribrats & MRI & segmentation & tif \\
tumorsimulations & MRI & segmentation & nii.gz \\
adenocarcima & CT & classification & npy \\
brain-organoids-segmentation-dataset & Microscopy & segmentation & tif \\
BraTS2020\_\allowbreak{}TrainingValidation & MRI & segmentation & nii \\
BraTS2020\_\allowbreak{}training\_\allowbreak{}data & MRI & segmentation & h5 \\
breast-cancer-cell-segmentation & Microscopy & segmentation & tif \\
SIPaKMeD & Microscopy & classification & bmp \\
computed-tomography-ct-of-the-brain & CT & classification & dcm \\
DMID-Breast-Cancer-Mammography & Mammography & segmentation & tif \\
DRIVE & Fundus & segmentation & tif \\
echonet-pediatric & Ultrasound & classification & avi \\
electron-microscopy-dataset & Microscopy & segmentation & tif \\
finding-lungs-in-ct-data & CT & segmentation & tif \\
fluorescence-microscopy-image-denoising-dataset & Microscopy & segmentation & h5 \\
fuse-my-cells-part-01 & Microscopy & classification & tif \\
fuse-my-cells-part-02 & Microscopy & classification & tif \\
kidney-nori-annotated-images & Microscopy & segmentation & tif \\
C-NMC\_\allowbreak{}Leukemia & Microscopy & classification & bmp \\
malignant-lymphoma-classification & Microscopy & classification & tif \\
MIAS-Mammography & Mammography & classification & pgm \\
mosmed-covid19-ct-scans & CT & classification & nii \\
TheChineseMammographyDatabase & Mammography & classification & dcm \\
PROSTATE\_\allowbreak{}MRI & MRI & classification & dcm \\
BTCV-TCIA-MultiOrgan-Abdominal & CT & segmentation & nii.gz \\
Comprehensive-3D-Medical-Dataset & CT & segmentation & nii.gz \\
Lung-PET-CT-Dx & CT & detection & dcm \\
Tibial-Plateau-Fracture & CT & classification & mat \\
\bottomrule
\end{tabular}
\caption{\label{tab:dataset_catalog_part2}
Complete dataset catalog used in the proposed benchmark (Part 2 of 3).
}
\end{table*}

\begin{table*}[t]
\centering
\footnotesize
\setlength{\tabcolsep}{3.0pt}
\renewcommand{\arraystretch}{0.98}
\begin{tabular}{@{}p{0.54\textwidth}p{0.19\textwidth}p{0.14\textwidth}p{0.08\textwidth}@{}}
\toprule
\textbf{Dataset name} & \textbf{Modality} & \textbf{Task type} & \textbf{Format} \\
\midrule
Brain-MP2RAGE-MPRAGE & MRI & segmentation & nii.gz \\
CHAOS-Abdominal & CT & segmentation & dcm \\
LNDb-Lung-Nodule & CT & segmentation & mhd \\
Totalsegmentator & CT & segmentation & nii.gz \\
L1-L5-Spine-FineSeg & CT & segmentation & nii.gz \\
HaN-Seg & CT & segmentation & nrrd \\
MoNuSAC-Cell-Detection & Microscopy & detection & tif \\
FLARE22 & CT & segmentation & nii.gz \\
Abdominal-CT-Scans & CT & segmentation & nii \\
DICOM-Brain-MRI & MRI & classification & dcm \\
RSNA-Heart-Detection & X-ray & detection & dcm \\
Hippocampus & MRI & segmentation & nii \\
KiTS23 & CT & segmentation & nii.gz \\
LiTS-Liver-Tumor-Segmentation & CT & segmentation & nii \\
LUNA16 & CT & detection & mhd \\
M3D-RefSeg & CT & segmentation & nii.gz \\
Prostate-MSD & MRI & segmentation & nii \\
SIIM-Medical-Images & CT & classification & dcm \\
Breast-MRI-NACT-Pilot & MRI & segmentation & dcm \\
CPTAC-CCRCC & CT+MRI & segmentation & dcm \\
CPTAC-PDA & CT+MRI & segmentation & dcm \\
CPTAC-UCEC & CT+MRI+PET & segmentation & dcm \\
LIDC-IDRI & CT & segmentation & dcm \\
Lumbar\_\allowbreak{}Spine\_\allowbreak{}Segmentation & MRI & segmentation & mha \\
MIDRC-RICORD-1A & CT & classification & dcm \\
NSCLC-Radiomics & CT & segmentation & dcm \\
QIN-BREAST & CT+MRI+PET & classification & dcm \\
Spine-Mets-CT-SEG & CT & segmentation & dcm \\
rsna-pneumonia-detection-challenge & X-ray & detection & dcm \\
soft\_\allowbreak{}tissue\_\allowbreak{}sarcoma & CT+MRI+PET & segmentation & dcm \\
tcga-luad & CT+PET & classification & dcm \\
LCTSC & CT & segmentation & dcm \\
\bottomrule
\end{tabular}
\caption{\label{tab:dataset_catalog_part3}
Complete dataset catalog used in the proposed benchmark (Part 3 of 3).
}
\end{table*}

\section{Cross-Group Generalization Analysis}
\label{app:cross_group}

We provide additional results by task type and medical modality.
This analysis examines whether model rankings remain stable across different subsets of the benchmark.
For each model $m$ and dataset $d$, we first compute the simple average over the eleven metrics:
\begin{equation}
\mathrm{Overall}(m,d)=\frac{1}{11}\sum_{k=1}^{11} v_k(m,d),
\end{equation}
where $v_k(m,d)$ denotes the score of metric $k$.
For a group $g$ with dataset set $\mathcal{D}_g$, the group score is:
\begin{equation}
\mathrm{Group}(m,g)=
\frac{1}{|\mathcal{D}_g|}
\sum_{d \in \mathcal{D}_g}
\mathrm{Overall}(m,d).
\end{equation}
All values are reported as percentages.
Groups with only three datasets are included for completeness and should be interpreted as descriptive evidence.

\subsection{Results by Medical Modality}

Table~\ref{tab:cross_modality_score} reports model performance across medical modality groups.
Gemini 3 Flash achieves the highest average score overall and ranks first in CT, MRI, Ultrasound, Ophthalmic Imaging, and X-ray.
Claude Opus 4.6 performs best on Microscopy / Pathology, while Claude Sonnet 4.6 performs best on Endoscopy and Composer 1.5 performs best on Mammography.
These results show that the strongest model overall is not always the best model for every medical modality.
Table~\ref{tab:cross_modality_rank} further summarizes cross-modality rank stability, showing that Gemini 3 Flash achieves the most stable overall ranking across modalities, although several models remain competitive on specific medical modality groups in the benchmark.

\begin{table*}[t]
\centering
\footnotesize
\setlength{\tabcolsep}{3pt}
\renewcommand{\arraystretch}{1.08}
\resizebox{\textwidth}{!}{
\begin{tabular}{lccccccccc}
\toprule
Model
& CT
& Microscopy / Path.
& MRI
& Ultrasound
& Ophthalmic
& Endoscopy
& X-ray
& Mammo.
& Mean \\
&
(\%)
& (\%)
& (\%)
& (\%)
& (\%)
& (\%)
& (\%)
& (\%)
& (\%) \\
\midrule

Gemini 3 Flash~\cite{googledeepmind2025gemini3flash}
& \textbf{65.5}
& 59.5
& \textbf{64.3}
& \textbf{72.6}
& \textbf{67.2}
& 59.8
& \textbf{66.4}
& 55.8
& \textbf{63.9} \\

Claude Opus 4.6~\cite{anthropic2026claudeopus46}
& 55.2
& \textbf{63.4}
& 51.2
& 61.2
& 61.8
& 59.7
& 48.5
& 57.7
& 57.3 \\

Claude Sonnet 4.6~\cite{anthropic2026claudesonnet46}
& 52.9
& 59.4
& 55.3
& 48.5
& 63.2
& \textbf{62.1}
& 65.3
& 35.8
& 55.3 \\

Composer 1.5~\cite{cursor2026composer15}
& 53.8
& 51.8
& 48.0
& 43.9
& 62.4
& 52.1
& 54.3
& \textbf{62.9}
& 53.6 \\

GPT-5.2-Codex~\cite{openai2025gpt52codex}
& 51.0
& 53.1
& 51.5
& 52.9
& 64.4
& 53.5
& 46.2
& 41.5
& 51.8 \\

GPT-5.2~\cite{openai2025gpt52}
& 50.4
& 56.7
& 48.3
& 50.6
& 53.3
& 46.8
& 55.2
& 41.7
& 50.4 \\

Kimi K2.5~\cite{kimiteam2026kimik25}
& 55.6
& 55.8
& 40.4
& 32.5
& 57.0
& 56.0
& 48.3
& 37.8
& 47.9 \\

Claude Haiku 4.5~\cite{anthropic2025claudehaiku45}
& 49.9
& 46.1
& 41.9
& 40.7
& 50.2
& 37.1
& 42.8
& 53.8
& 45.3 \\

Grok 4.20~\cite{xai2026grok420}
& 55.4
& 46.8
& 42.5
& 29.6
& 48.8
& 35.4
& 50.1
& 39.3
& 43.5 \\

\bottomrule
\end{tabular}
}
\caption{Cross-modality results measured by the simple average of the eleven metrics. Bold numbers indicate the best model in each group.}
\label{tab:cross_modality_score}
\end{table*}

\begin{table*}[t]
\centering
\small
\setlength{\tabcolsep}{1.35pt}
\renewcommand{\arraystretch}{1.08}

\begin{tabular}{@{}lccccccccccc@{}}
\toprule
Model & Avg. Rank & Std. & Stability & CT & Mic./Path. & MRI & US & Ophth. & Endo. & X-ray & Mammo. \\
\midrule

Gemini 3 Flash~\cite{googledeepmind2025gemini3flash}
& \textbf{1.5} & \textbf{0.76} & Stable
& 1 & 2 & 1 & 1 & 1 & 2 & 1 & 3 \\

Claude Opus 4.6~\cite{anthropic2026claudeopus46}
& 3.4 & 1.69 & Moderate
& 4 & 1 & 4 & 2 & 5 & 3 & 6 & 2 \\

Claude Sonnet 4.6~\cite{anthropic2026claudesonnet46}
& 3.9 & 2.64 & Unstable
& 6 & 3 & 2 & 5 & 3 & 1 & 2 & 9 \\

Composer 1.5~\cite{cursor2026composer15}
& 4.9 & 1.89 & Moderate
& 5 & 7 & 6 & 6 & 4 & 6 & 4 & 1 \\

GPT-5.2-Codex~\cite{openai2025gpt52codex}
& 5.0 & 2.14 & Moderate
& 7 & 6 & 3 & 3 & 2 & 5 & 8 & 6 \\

GPT-5.2~\cite{openai2025gpt52}
& 5.4 & 1.77 & Moderate
& 8 & 4 & 5 & 4 & 7 & 7 & 3 & 5 \\

Kimi K2.5~\cite{kimiteam2026kimik25}
& 6.1 & 2.36 & Moderate
& 2 & 5 & 9 & 8 & 6 & 4 & 7 & 8 \\

Grok 4.20~\cite{xai2026grok420}
& 7.1 & 2.17 & Moderate
& 3 & 8 & 7 & 9 & 9 & 9 & 5 & 7 \\

Claude Haiku 4.5~\cite{anthropic2025claudehaiku45}
& 7.8 & 1.67 & Moderate
& 9 & 9 & 8 & 7 & 8 & 8 & 9 & 4 \\

\bottomrule
\end{tabular}

\caption{Cross-modality rank stability. Lower average rank and lower standard deviation indicate more stable performance across modalities.}
\label{tab:cross_modality_rank}
\end{table*}

\subsection{Results by Task Type}

Table~\ref{tab:cross_task_score} reports results by task type.
Gemini 3 Flash ranks first on classification, segmentation, and detection, showing the most stable task-level generalization.
Claude Opus 4.6 and Claude Sonnet 4.6 remain strong on classification and segmentation, while the detection group shows larger ranking changes.
Because detection contains only six datasets, these results should be treated as indicative rather than definitive.
Table~\ref{tab:cross_task_rank} further reports task-level rank stability, confirming that Gemini 3 Flash is the only model ranked first across all three task types, while other models show larger rank variation, especially on detection.

\begin{table}[t]
\centering
\small
\setlength{\tabcolsep}{1.2pt}
\renewcommand{\arraystretch}{1.05}

\scalebox{0.96}{
\begin{tabular}{@{}lcccc@{}}
\toprule
Model & Seg. & Cls. & Det. & Mean \\
      & (\%) & (\%) & (\%) & (\%) \\
\midrule

Gemini 3 Flash~\cite{googledeepmind2025gemini3flash}
& \textbf{62.6} & \textbf{63.0} & \textbf{74.3} & \textbf{66.6} \\

Claude Sonnet 4.6~\cite{anthropic2026claudesonnet46}
& 54.8 & 56.6 & 59.9 & 57.1 \\

Claude Opus 4.6~\cite{anthropic2026claudeopus46}
& 56.7 & 57.7 & 54.9 & 56.4 \\

Composer 1.5~\cite{cursor2026composer15}
& 49.5 & 55.4 & 56.0 & 53.6 \\

GPT-5.2~\cite{openai2025gpt52}
& 50.5 & 51.8 & 54.3 & 52.2 \\

GPT-5.2-Codex~\cite{openai2025gpt52codex}
& 54.1 & 49.8 & 51.2 & 51.7 \\

Kimi K2.5~\cite{kimiteam2026kimik25}
& 54.7 & 47.1 & 49.4 & 50.4 \\

Grok 4.20~\cite{xai2026grok420}
& 49.0 & 44.7 & 54.3 & 49.3 \\

Claude Haiku 4.5~\cite{anthropic2025claudehaiku45}
& 49.3 & 41.2 & 53.3 & 47.9 \\

\bottomrule
\end{tabular}
}

\caption{Cross-task results measured by the simple average of the eleven metrics.}
\label{tab:cross_task_score}
\end{table}

\begin{table}[t]
\centering
\small
\setlength{\tabcolsep}{1.0pt}
\renewcommand{\arraystretch}{1.05}

\resizebox{1\columnwidth}{!}{
\begin{tabular}{@{}lcccccc@{}}
\toprule
Model & Avg. & Std. & Stab. & Seg. & Cls. & Det. \\
\midrule

Gemini 3 Flash~\cite{googledeepmind2025gemini3flash}
& \textbf{1.0} & \textbf{0.00} & Stable
& 1 & 1 & 1 \\

Claude Opus 4.6~\cite{anthropic2026claudeopus46}
& 2.7 & 1.15 & Stable
& 2 & 2 & 4 \\

Claude Sonnet 4.6~\cite{anthropic2026claudesonnet46}
& 2.7 & 0.58 & Stable
& 3 & 3 & 2 \\

Composer 1.5~\cite{cursor2026composer15}
& 4.7 & 2.08 & Moderate
& 7 & 4 & 3 \\

GPT-5.2~\cite{openai2025gpt52}
& 5.7 & 0.58 & Stable
& 6 & 5 & 6 \\

GPT-5.2-Codex~\cite{openai2025gpt52codex}
& 6.3 & 1.53 & Moderate
& 5 & 6 & 8 \\

Kimi K2.5~\cite{kimiteam2026kimik25}
& 6.7 & 2.52 & Unstable
& 4 & 7 & 9 \\

Grok 4.20~\cite{xai2026grok420}
& 7.3 & 2.08 & Moderate
& 9 & 8 & 5 \\

Claude Haiku 4.5~\cite{anthropic2025claudehaiku45}
& 8.0 & 1.00 & Stable
& 8 & 9 & 7 \\

\bottomrule
\end{tabular}
}

\caption{Cross-task rank stability. Gemini 3 Flash is the only model ranked first across all three task types.}
\label{tab:cross_task_rank}
\end{table}

\subsection{Summary}

The cross-group analysis shows that Gemini 3 Flash is the most robust model across both task types and medical modalities.
It ranks first on all three task types and has the best average modality rank.
However, model performance is not fully invariant across data conditions.
Claude Opus 4.6 performs best on Microscopy / Pathology, Claude Sonnet 4.6 performs best on Endoscopy, and Composer 1.5 performs best on Mammography.
These results suggest that raw medical data standardization difficulty depends not only on the target task, but also on the medical modality and the conventions of the underlying data source.

\section{Evaluation Details}
\label{app:evaluation_details}
This appendix provides the full definition of the eleven-metric evaluation protocol used in the main experiments. The protocol is designed to evaluate whether a model can transform raw heterogeneous medical data into source-grounded standardized images, per-image annotations, and dataset annotations. It therefore checks not only output format, but also source selection, semantic alignment, information quality, content fidelity, and dataset-level metadata quality.

\subsection{Notation}
\label{app:notation}

For each dataset, let $N$ denote the number of target samples specified by the sample manifest. For sample $i$, the evaluator records the following quantities:
\begin{itemize}
    \setlength{\itemsep}{1pt}
    \setlength{\parsep}{0pt}
    \setlength{\parskip}{0pt}
    \setlength{\topsep}{2pt}
    \item $p_i \in \{0,1\}$: whether the sample forms a valid image-text pair. This requires a readable image, a parseable JSON annotation, and a JSON image path that matches the actual output image.
    \item $s_i \in \{0,1\}$: source matching score between the predicted source information and the verified raw source file or source record.
    \item $g_i \in [0,1]$: schema-validity score, computed as the fraction of schema checks passed.
    \item $q_i \in [0,1]$: semantic correctness score for sample-level fields.
    \item $c_i \in [0,1]$: information completeness score.
    \item $d_i \in [0,1]$: information validity score.
    \item $r_i \in [0,1]$: information non-redundancy score.
    \item $f_i \in [0,1]$: content fidelity score.
\end{itemize}

We define the set of source-matched samples as
\begin{equation}
S_{\mathrm{src}}=\{i\mid s_i>0\}, \qquad
N_{\mathrm{src}}=|S_{\mathrm{src}}|.
\end{equation}
Field-level metrics that require a meaningful comparison with the target ground truth are computed over $S_{\mathrm{src}}$. Joint metrics are computed over all $N$ target samples, so wrong-source and missing-output cases are penalized directly.

\subsection{Output Validity and Source Matching}
\label{app:output_source}

The evaluator first verifies whether each sample has a valid paired output. A pair is considered valid only when the standardized image exists, can be opened as an image file, the JSON annotation exists and can be parsed, and the image path recorded in the JSON points to the actual standardized image. This check produces $p_i$.
Let $I_i^{\mathrm{read}}$, $J_i^{\mathrm{parse}}$, and
$P_i^{\mathrm{match}}$ indicate whether the standardized image can be read, the JSON can be parsed, and the JSON image path matches the standardized image, respectively. We define
\begin{equation}
p_i =
\mathbb{I}\!\left[
I_i^{\mathrm{read}}
\land J_i^{\mathrm{parse}}
\land P_i^{\mathrm{match}}
\right].
\end{equation}

Source matching evaluates whether the predicted source fields can be traced to the verified raw source. The evaluator normalizes source identifiers by handling common path, filename, extension, and stem variations. Let $A_i^{\mathrm{pred}}$ and $A_i^{\mathrm{gt}}$ denote the normalized predicted and verified source-alias sets, respectively, and let $t_i\in\{0,1\}$ indicate whether explicit source-trace evidence is present. We define
\begin{equation}
s_i=\mathbb{I}\left[
A_i^{\mathrm{pred}}\cap A_i^{\mathrm{gt}}\neq\emptyset
\land t_i=1
\right].
\end{equation}
Thus, $s_i=1$ only when the predicted source matches a verified source alias and is supported by explicit trace evidence; otherwise, $s_i=0$. This check is used as a gate for field-level semantic and content evaluation, because evaluating annotations against the target sample is not meaningful when the model has selected an unrelated source.

\subsection{Structure Metrics}
\label{app:structure_metrics}

\paragraph{Schema Validity (SV).}
SV measures whether the sample-level JSON follows the required schema. The evaluator checks required top-level blocks, record fields, image-path references, context fields, media fields, task-specific branches, spatial branches, and branch exclusivity. Let $g_i$ be the fraction of schema checks passed for sample $i$.
Specifically, let $C_{ik}\in\{0,1\}$ indicate whether sample $i$ passes the $k$-th schema check. Since the evaluator contains 14 fixed schema checks,
\begin{equation}
g_i = \frac{1}{14}\sum_{k=1}^{14} C_{ik}.
\end{equation}
Thus, the default threshold $g_i\geq0.85$ requires at least 12 of the 14 schema checks to pass. The dataset-level schema validity score is then
\begin{equation}
\mathrm{SV}=\frac{1}{N}\sum_{i=1}^{N} g_i .
\end{equation}

\paragraph{Schema-Semantic Composite (SSC).}
SV alone can be high when a model generates a well-formed but semantically incorrect JSON. SSC therefore couples schema validity with semantic correctness:
\begin{equation}
\mathrm{SSC}=\frac{1}{N}\sum_{i=1}^{N} g_i q_i .
\end{equation}
This metric penalizes outputs that satisfy the surface schema but assign incorrect task type, modality, dimension, dataset identity, or annotation semantics for the sample.

\subsection{Semantic Metric}
\label{app:semantic_metric}

\paragraph{Semantic Correctness (SC).}
SC evaluates whether key sample-level fields match the verified ground truth. These fields include dataset identity, task type, modality, data dimension, class information, and annotation type. 
For sample $i$, let $J_i$ denote the set of applicable semantic fields, $w_j$ the weight of field $j$, and $m_j(i)\in\{0,1\}$ the matching result after normalization. We first compute
\begin{equation}
S_{\mathrm{fields}}(i)=
\frac{\sum_{j\in J_i} w_j m_j(i)}
{\sum_{j\in J_i} w_j}.
\end{equation}
The sample-level semantic correctness score is then
\begin{equation}
\begin{aligned}
q_i ={}&
0.70S_{\mathrm{fields}}(i)
+0.10S_{\mathrm{presence}}(i) \\
&+0.20S_{\mathrm{task}}(i).
\end{aligned}
\end{equation}
where $S_{\mathrm{fields}}(i)$ measures agreement of applicable semantic fields, $S_{\mathrm{presence}}(i)$ measures required-field presence, and $S_{\mathrm{task}}(i)$ measures task-specific consistency.
Before normalization over the applicable fields, the evaluator assigns weights of $0.25$ to task type, $0.15$ to data dimension, $0.20$ to modality, $0.15$ to dataset identity, $0.15$ to class information, and $0.10$ to annotation or map type. Fields without a verified target value are excluded, and the remaining weights are renormalized in $S_{\mathrm{fields}}(i)$.

$S_{\mathrm{presence}}(i)$ is the mean of task-dependent binary presence checks. For classification, these checks cover the label and its textual class description. For detection and segmentation, they cover the schema variant, class information, and the required box or mask payload. $S_{\mathrm{task}}(i)$ is the mean field-level agreement between the generated and verified task-payload blocks, using the deterministic matching rules described below. When no verified task-payload block is available, it defaults to $S_{\mathrm{presence}}(i)$. Since these fields are meaningful only after a related source is identified, SC is computed over the source-matched set:
\begin{equation}
\mathrm{SC}=
\frac{1}{N_{\mathrm{src}}}
\sum_{i\in S_{\mathrm{src}}} q_i .
\end{equation}
If no sample passes the source gate, the score is set to 0.

\subsection{Information and Fidelity Metrics}
\label{app:information_metrics}

\paragraph{Information Completeness (IC).}
IC measures whether required auxiliary information is provided for a sample. It checks the presence of the applicable expected auxiliary field groups and summarizes these checks into a sample-level completeness score $c_i$. We then aggregate $c_i$ over the source-matched samples:
\begin{equation}
\mathrm{IC}=
\frac{1}{N_{\mathrm{src}}}
\sum_{i\in S_{\mathrm{src}}} c_i .
\end{equation}

\paragraph{Information Validity (IV).}
IV measures whether populated auxiliary fields contain meaningful non-placeholder values. It checks whether the applicable expected auxiliary field groups contain valid values rather than empty tokens, placeholders, or invalid default values, yielding a sample-level validity score $d_i$. The IV score is computed as:
\begin{equation}
\mathrm{IV}=
\frac{1}{N_{\mathrm{src}}}
\sum_{i\in S_{\mathrm{src}}} d_i .
\end{equation}

\paragraph{Information Non-Redundancy (INR).}
INR measures whether the output avoids repetitive or redundant field filling. It rewards outputs that provide compact, non-duplicative information while retaining meaningful expected fields, yielding a sample-level non-redundancy score $r_i$:
\begin{equation}
\mathrm{INR}=
\frac{1}{N_{\mathrm{src}}}
\sum_{i\in S_{\mathrm{src}}} r_i .
\end{equation}

\paragraph{Content Fidelity (CF).}
CF measures whether recovered task-specific content matches the verified ground truth. This includes labels, class names, masks, boxes, source mappings, expected auxiliary fields, and task-specific annotation payloads:
For sample $i$, let $B_i$ denote the set of applicable expected content blocks and let $F_{ib}$ denote the verified fields in block $b$. We compute the sample-level content-fidelity score as
\begin{equation}
\label{eq:content_fidelity}
f_i =
\frac{1}{|B_i|}
\sum_{b\in B_i}
\left(
\frac{1}{|F_{ib}|}
\sum_{j\in F_{ib}}\phi_{ibj}
\right),
\end{equation}
where $\phi_{ibj}\in[0,1]$ is the deterministic agreement score between the generated and verified values of field $j$.

Numerical and Boolean fields use exact matching. Text and enumerated fields use normalized exact matching. Path fields use normalized full-path or stem-alias matching according to the ground-truth specification. For list-valued fields, the evaluator computes Jaccard overlap between the normalized predicted and verified sets:
\begin{equation}
\phi_{\mathrm{list}} =
\frac{|A^{\mathrm{pred}}\cap A^{\mathrm{gt}}|}
{|A^{\mathrm{pred}}\cup A^{\mathrm{gt}}|}.
\end{equation}
Nested JSON objects are compared recursively and averaged over their verified subfields. When the verified value is \texttt{null}, the generated field receives credit only if it is also effectively empty. None of these rules uses embedding-based or cosine similarity. All main experiments use arithmetic averaging across the applicable content blocks, as defined in Eq.~\eqref{eq:content_fidelity}.

\begin{equation}
\mathrm{CF}=
\frac{1}{N_{\mathrm{src}}}
\sum_{i\in S_{\mathrm{src}}} f_i .
\end{equation}
Together, IC, IV, INR, and CF distinguish evidence-supported content recovery from merely producing a syntactically valid JSON file.

\begin{table*}[t]
\centering
\small
\setlength{\tabcolsep}{5pt}
\resizebox{\textwidth}{!}{%
\begin{tabular}{@{}lll@{\hspace{1.8em}}lcc@{}}
\toprule
\multicolumn{3}{c}{\textbf{Threshold sensitivity}} &
\multicolumn{3}{c}{\textbf{Model-ranking stability}} \\
\cmidrule(lr){1-3}\cmidrule(lr){4-6}
\textbf{Varied} &
\textbf{Fixed thresholds} &
\textbf{Mean E2E (\%)} &
\textbf{Setting} &
\textbf{Thresholds $(\tau_g,\tau_q,\tau_f)$} &
\textbf{Consistency} \\
\midrule
$g_i$ &
$q_i=0.50,\ f_i=0.50$ &
$0.75$: 45.50,\ $0.85$: 30.15,\ $1.00$: 8.11 &
Loose &
$(0.80,0.40,0.40)$ &
31/36 (86.1\%) \\

$q_i$ &
$g_i=0.85,\ f_i=0.50$ &
$0.40$: 30.56,\ $0.50$: 30.15,\ $0.60$: 27.99 &
Default &
$(0.85,0.50,0.50)$ &
-- \\

$f_i$ &
$g_i=0.85,\ q_i=0.50$ &
$0.40$: 36.77,\ $0.50$: 30.15,\ $0.60$: 20.22 &
Strict &
$(0.90,0.60,0.60)$ &
28/36 (77.8\%) \\
\bottomrule
\end{tabular}%
}
\caption{Threshold sensitivity and model-ranking stability. Mean E2E is averaged across the nine models after dataset-level macro-averaging. Pairwise consistency under the loose and strict settings is measured relative to the default setting.}
\label{tab:threshold_analysis}
\end{table*}
\subsection{Metadata Metrics}
\label{app:metadata_metrics}

Each dataset also contains a dataset-level metadata record. The evaluator checks whether this record correctly summarizes global dataset information, including dataset name, primary modality, task type, data volume, valid-image count, and class information.

\paragraph{Metadata Semantic Correctness (MSC).}
MSC measures whether dataset-level semantic fields match the verified dataset-level ground truth:
\begin{equation}
\mathrm{MSC}=m_{\mathrm{sc}},
\end{equation}
where $m_{\mathrm{sc}}$ denotes the metadata semantic correctness score.

\paragraph{Meta-Sample Joint Score (MSJ).}
Dataset-level metadata may be correct even when sample-level outputs are unfaithful. We therefore define a joint metadata score that couples metadata reliability with sample-level content fidelity:
\begin{equation}
\mathrm{MSJ}
=
(\tilde{m}_{sv}\,\tilde{m}_{sc}\,\tilde{m}_{cf})^{1/3}
\cdot \mathrm{CF},
\end{equation}
where $\tilde{m}_{x}=\max(m_x,10^{-6})$, and
$m_{sv}$, $m_{sc}$, and $m_{cf}$ denote metadata schema validity, metadata semantic correctness, and metadata content fidelity, respectively. The geometric mean penalizes metadata records with any weak component, and multiplication by CF further requires consistency between dataset-level metadata and sample-level standardized content.

\subsection{Joint Metrics}
\label{app:joint_metrics}

\paragraph{Source Content Joint Score (SCJ).}
SCJ evaluates whether source selection and content preservation succeed simultaneously:
\begin{equation}
\mathrm{SCJ}=
\frac{1}{N}\sum_{i=1}^{N} s_i f_i .
\end{equation}
Unlike CF, which is computed only after source matching, SCJ is averaged over all target samples. Thus, a wrong-source output receives no credit even if its generated content appears plausible.

\paragraph{End-to-End Strict Pass (E2E).}
E2E measures whether a sample passes the sample-level standardization pipeline.
We first define the strict pass indicator:
\begin{equation}
\begin{aligned}
e_i = \mathbb{I}\big[
& p_i \land s_i = 1 \land g_i \ge 0.85 \\
& \land q_i \ge 0.5 \land f_i \ge 0.5
\big].
\end{aligned}
\end{equation}
The E2E score is then
\begin{equation}
\mathrm{E2E} = \frac{1}{N}\sum_{i=1}^{N} e_i .
\end{equation}
A sample passes E2E only when it has a valid image-text pair, selects the exact source, reaches the schema-validity threshold, and satisfies the minimum semantic and content-fidelity thresholds.

\subsection{Threshold Sensitivity Analysis}
\label{app:threshold_analysis}

The default E2E thresholds are selected through preliminary experiments based on manual checks of whether the resulting standardized data remain usable for downstream tasks. We examine the sensitivity of E2E to each threshold while fixing the other two at their default values. We further evaluate model-ranking stability under loose, default, and strict threshold settings.

As shown in Table~\ref{tab:threshold_analysis}, changing the thresholds affects the absolute E2E score, as expected. Around the default setting, changing $q_i$ produces relatively small differences, whereas changing $f_i$ has a larger effect, indicating greater sensitivity to content fidelity than to semantic correctness. Among the 36 pairwise model comparisons, the loose and strict settings preserve 31 and 28 orderings relative to the default setting, respectively. Gemini 3 Flash remains the top-ranked E2E model under both settings, although several middle-ranked models exchange positions. Thus, the main conclusion remains stable, while fine-grained rankings are moderately sensitive to threshold choices.

\subsection{Dataset-Level Aggregation}
\label{app:aggregation}

All eleven metrics are first computed independently for each dataset. We then average each metric across datasets with equal weight:
\begin{equation}
\mathrm{Score}(k)=\frac{1}{|\mathcal{D}|}
\sum_{D\in\mathcal{D}} \mathrm{Score}_{D}(k),
\end{equation}
where $\mathcal{D}$ is the set of 100 benchmark datasets and $k$ denotes one of the eleven metrics. This prevents large datasets from dominating the final results and makes the reported scores reflect performance across heterogeneous medical domains, modalities, source formats, annotation styles, and directory organizations.

\subsection{Metric Groups Used in Analysis}
\label{app:metric_groups}

For analysis, we organize the eleven metrics into five capability groups:
\begin{itemize}
    \setlength{\itemsep}{-0.5pt}
    \setlength{\parsep}{0pt}
    \setlength{\parskip}{0pt}
    \item \textbf{Structure}: SV and SSC.
    \item \textbf{Semantic}: SC.
    \item \textbf{Content}: IC, IV, INR, and CF.
    \item \textbf{Metadata}: MSC and MSJ.
    \item \textbf{Joint}: SCJ and E2E.
\end{itemize}
These groups are used in the main result table and in error attribution analysis. They separate outputs that are merely well-formed from outputs that are source-grounded, semantically correct, content-faithful, and usable as standardized medical data.

\section{Agentic Setup}
\label{app:agentic_setup}
\paragraph{Prompting protocol.}
The prompt specifies the raw dataset directory, sample manifest, required output structure, standardized JSON schema, and task-specific branches. It instructs the agent to inspect raw files and documentation, identify source evidence, convert medical images, align annotations and metadata, and verify the generated outputs. Every populated field must be supported by source evidence, while unavailable information is set to null.

\paragraph{Tool access.}
The agent operates in a local workspace with file-system access to inspect dataset directories, read raw files, parse metadata and annotation files, execute scripts for format conversion or validation, and write standardized image and JSON outputs. It cannot access ground-truth annotations, evaluation scores, or manual feedback during generation.

\paragraph{Agent steps and token budget.}
We do not impose a fixed number of agent steps or a fixed token budget per sample. Actual usage varies with the amount of source data, tool interaction, and generated output.

\paragraph{Retry and rate limiting.}
Retries are used only for infrastructure-level failures, such as interrupted runs or provider-side errors, and are not used to improve model scores. No additional rate limiting is applied beyond the default service limits.

\section{Inference Strategy Details}
\label{app:inference_strategies}

\begin{table}[t]
\centering
\small
\setlength{\tabcolsep}{2.5pt}
\resizebox{\columnwidth}{!}{
\begin{tabular}{lccccccccccc}
\toprule
\textbf{Strategy} & \textbf{SV} & \textbf{SSC} & \textbf{SC} & \textbf{IC} & \textbf{IV} & \textbf{INR} & \textbf{CF} & \textbf{MSC} & \textbf{MSJ} & \textbf{SCJ} & \textbf{E2E} \\
\midrule
S0 & 82.2 & 49.8 & 49.1 & 47.5 & 41.5 & 42.5 & 40.3 & 56.8 & 34.1 & 38.0 & 24.9 \\
S3 & 88.6 & 59.6 & 59.9 & 68.1 & 62.1 & 59.9 & 54.5 & 58.8 & 46.2 & 52.0 & 45.6 \\
\bottomrule
\end{tabular}
}
\caption{Comparison of S0 and S3 on Claude Haiku 4.5. All values are percentages (\%).}
\label{tab:haiku_s0_s3}
\end{table}
We compare five inference strategies, denoted as S0-S4. S0 serves as the direct-generation baseline. S1 and S2 refine the S0 output in different ways, while S3 and S4 use multiple independently generated candidates for selection.

\paragraph{Direct Generation (S0).}
S0 directly performs the standardization task. Given the raw dataset directory and sample manifest, the model identifies the relevant source files, converts medical data into standardized images, aligns annotations and metadata, and generates the required image-JSON outputs. No additional refinement or candidate selection is applied.

\paragraph{Self-Refinement (S1).}
S1 starts from the S0 output and asks the same model to review the generated result against the raw data, sample manifest, and target schema. The review covers source mapping, schema validity, semantic fields, annotations, and metadata. The model can revise incorrect fields or regenerate affected outputs when supported by the source evidence. Ground truth and evaluation results are not available during refinement.

\paragraph{Verification-Guided Refinement (S2).}
S2 also starts from the S0 output, but performs more restricted corrections. The model follows a predefined verification checklist and focuses on high-risk fields, including task type, dimension, active task branches, labels and class IDs, image paths, source information, annotation type, and metadata counts. Other fields remain unchanged. This strategy therefore limits the refinement to fields that are more likely to affect standardization quality.

\paragraph{Best-of-$n$ Selection (S3).}
S3 generates multiple complete outputs independently and selects one complete candidate for each sample. A deterministic validator evaluates each candidate using checks such as schema validity, image readability, image-JSON alignment, task and dimension consistency, source traceability, field completeness, and task-specific content validity. The candidate with the highest validation score is selected. The validator does not use ground truth or evaluation scores.

\paragraph{Field-Wise Selection (S4).}
S4 also starts from multiple candidate outputs. It first uses the same validator to identify the strongest candidate as the base output. It then compares candidates only for a limited set of high-risk fields and selects the most consistent values for these fields. The remaining fields are kept from the base candidate. Unlike S3, which selects one complete output, S4 combines selected fields across candidates.

\paragraph{Additional Ablation on Claude Haiku 4.5.}
The main ablation in Sec.~4.4 uses Gemini 3 Flash. To examine whether the benefit of S3 also holds for a weaker model, we further compare S0 and S3 on Claude Haiku 4.5. As shown in Table~\ref{tab:haiku_s0_s3}, S3 improves SCJ from 38.0\% to 52.0\% and E2E from 24.9\% to 45.6\%, corresponding to gains of 14.0 and 20.7 percentage points, respectively. Improvements are also observed across the remaining metrics. These results show that S3 also benefits a weaker model.

\end{document}